%% file: main.tex
\documentclass{article}

\usepackage{microtype}
\usepackage{graphicx}
\usepackage{subcaption}
\usepackage{booktabs} % for professional tables

\usepackage{hyperref}

\usepackage[accepted]{icml2020}

\icmltitlerunning{Revisiting Training Strategies and Generalization Performance in Deep Metric Learning}

\usepackage{times}
\usepackage{epsfig}
\usepackage{graphicx}
\usepackage{amsmath}
\usepackage{amssymb}
\usepackage[utf8]{inputenc} % allow utf-8 input
\usepackage{url}            % simple URL typesetting
\usepackage{booktabs}       % professional-quality tables
\usepackage{amsfonts}       % blackboard math symbols
\usepackage{nicefrac}       % compact symbols for 1/2, etc.
\usepackage{microtype}      % microtypography
\usepackage{color}
\usepackage[table]{xcolor}
\graphicspath{ {./images/} }
\usepackage{multirow}
\DeclareMathOperator*{\argmax}{arg\,max}
\DeclareMathOperator*{\argmin}{arg\,min}

\newcommand{\blue}[1]{\textcolor{blue}{#1}}

\begin{document}

\twocolumn[
\icmltitle{Revisiting Training Strategies and Generalization Performance \\ in Deep Metric Learning}


\icmlsetsymbol{equal}{*}

\begin{icmlauthorlist}
\icmlauthor{Karsten Roth}{equal,mila,hd}
\icmlauthor{Timo Milbich}{equal,hd}
\icmlauthor{Samarth Sinha}{mila,tor}
\icmlauthor{Prateek Gupta}{mila,ox}
\icmlauthor{Björn Ommer}{hd}
\icmlauthor{Joseph Paul Cohen}{mila}
\end{icmlauthorlist}

\icmlaffiliation{mila}{Mila, Université de Montréal}
\icmlaffiliation{hd}{HCI/IWR, Heidelberg University}
\icmlaffiliation{tor}{University of Toronto}
\icmlaffiliation{ox}{The Alan Turing Institute, University of Oxford}

\icmlcorrespondingauthor{Karsten Roth}{karsten.rh1@gmail.com}

\icmlkeywords{Machine Learning, ICML}

\vskip 0.3in
]



\printAffiliationsAndNotice{\icmlEqualContribution} % otherwise use the standard text.

\begin{abstract}
Deep Metric Learning (DML) is arguably one of the most influential lines of research for learning visual similarities with many proposed approaches every year. Although the field benefits from the rapid progress, the divergence in training protocols, architectures, and parameter choices make an unbiased comparison difficult. To provide a consistent reference point, we revisit the most widely used DML objective functions and conduct a study of the crucial parameter choices as well as the commonly neglected mini-batch sampling process. Under consistent comparison, DML objectives show much higher saturation than indicated by literature. Further based on our analysis, we uncover a correlation between the embedding space density and compression to the generalization performance of DML models. Exploiting these insights, we propose a simple, yet effective, training regularization to reliably boost the performance of ranking-based DML models on various standard benchmark datasets. Code and a publicly accessible WandB-repo are available at \url{https://github.com/Confusezius/Revisiting_Deep_Metric_Learning_PyTorch}.
% Deep Metric Learning (DML) is arguably one of the most influential lines of research for learning visual similarities with many proposed approaches every year. Although the field benefits from the rapid progress, the divergence in training protocols, architectures, and parameter choices make an unbiased comparison difficult. To provide a consistent reference point, we revisit the most widely used DML objective functions and conduct a study of the crucial parameter choices as well as the commonly neglected mini-batch sampling process. Based on our analysis, we uncover a correlation between the embedding space density and the generalization performance of DML models. Exploiting these insights, we propose a simple, yet effective, training regularization to reliably boost the performance of ranking-based DML models on various standard benchmark datasets.
\end{abstract}

%%%%%%%%%%%%%%%%%%%%%%%%%%%%%%%%%%%%%%%%%%%%%%%%%%%%%%%%%%%%%%%%%%%%%%%%%%%%%%%%%%%%%%%%%
\section{Introduction}

\input{figures/first_page.tex}

Learning visual similarity is important for a wide range of vision tasks, such as image clustering \cite{grouping}, face detection \cite{semihard} or image retrieval \cite{margin}. Measuring similarity requires learning an embedding space which captures images and reasonably reflects similarities using a defined distance metric. One of the most adopted classes of algorithms for this task is Deep Metric Learning (DML) which leverages deep neural networks to learn such a distance preserving embedding.\\
Due to the growing interest in DML, a large corpus of literature has been proposed contributing to its success. However, as recent DML approaches explore more diverse research directions such as architectures~\cite{dreml,horde}, objectives functions~\cite{rankedlist,signal2noise} and additional training tasks~\cite{mic,dvml}, an unbiased comparison of their results becomes more and more difficult.
%Due to the success and growing interest in DML, many approaches are proposed every year. Despite greatly benefiting from such rapid progress, the divergence in training protocols, architectures, and parameter choices makes an unbiased comparison of their results difficult. 
Further, undisclosed technical details (s.a. data augmentations or training regularization) pose a challenge to the reproducibility of such methods, which is of great concern in the machine learning community in general \citep{bouthillier2019unreproducible}. One goal of this work is to counteract this worrying trend by providing a comprehensive comparison of important and current DML baselines under identical training conditions on standard benchmark datasets (Fig. \ref{fig:fp1}). In addition, we thoroughly review common design choices of DML models which strongly influence generalization performance to allow for better comparability of current and future work.\\
On that basis, we extend our analysis to: \textit{(i)} The process of data sampling which is well-known to impact the DML optimization \cite{semihard}. While previous works only studied this process in the specific context of triplet mining strategies for ranking-based objectives \cite{margin, smartmining}, we examine the model-agnostic case of sampling informative mini-batches. \textit{(ii)} The generalization capabilities of DML models by analyzing the structure of their learned embedding spaces. While we are not able to reliably link typically targeted concepts such as large inter-class margins~\cite{sphereface,arcface} and intra-class variance~\cite{dvml} to generalization performance, we uncover a strong correlation to the compression of the learned representations.
%To this end, we examine intra-class-distances and inter-class-distances which have recently gained interest in the community \cite{mic, dvml}. 
% Our findings uncover that these do not provide a reliable metric to predict generalization performance across different datasets and methods. 
% We uncover a strong correlation between that the singular value spectrum of a learned embedding space strongly correlates with generalization abilities of a DML model. Rather, we show that the singular value spectrum of a learned embedding space strongly correlates with generalization abilities of a DML model. 
Lastly, based on this observation, we propose a simple, yet effective, regularization technique which effectively boosts the performance of ranking-based approaches on standard benchmark datasets as also demonstrated in Fig.~\ref{fig:fp1}. In summary, our most important contributions can be described as follows:
\begin{itemize}
    \item We provide an exhaustive analysis of recent DML objective functions, their training strategies, the influence of data-sampling, and model design choices to set a standard benchmark. To this end, we will make our code publicly available.

    \item We provide new insights into DML generalization by analyzing its correlation to the embedding space compression (as measured by its spectral decay), inter-class margins and intra-class variance.
    
    \item Based on the result above, we propose a simple technique to regularize the embedding space compression which we find to boost generalization performance of ranking-based DML approaches.
\end{itemize}

This work is structured as follows: After reviewing related work in \S \ref{sec:related_work}, we discuss and motivate our analyzed components of DML models and their training setup in \S \ref{sec:dml}. Finally in \S \ref{sec:eval_ana} we present the findings of our study, analyze DML generalization in \S \ref{sec:generalization} and close with a conclusion in \S \ref{sec:conclusion}.

%%%%%%%%%%%%%%%%%%%%%%%%%%%%%%%%%%%%%%%%%%%%%%%%%%%%%%%%%%%%%%%%%%%%%%%%%%%%%%%%%%%%%%%%%
%\paragraph{Knowledge Distillation \& Coreset Selection}
%Knowledge distillation and information compression has been widely studied. For Deep Learning, primary research has gone into distilling information encoded in neural networks by learning smaller student networks\cite{knowledgedistill} or activation compression\cite{fishergan} through coreset selection. Our work however examines batch compression for Deep Metric Learning and more closely follows the idea of beneficial subset selection as was proposed in \cite{wei_markov} for hidden Markov models in an unsupervised way or \cite{savareseactive} for batch-mode active learning. In contrast, we specifically examine batch subset selection on a continuously updated memory bank for supervised Deep Metric Learning.
% as an alernative to the previous paragraph

\section{Related Works}
\label{sec:related_work}

\textbf{Deep Metric Learning:}
Deep Metric Learning (DML) has become increasingly important for applications ranging from image retrieval \cite{proxynca,mic,margin,dvml} to zero-shot classification \cite{semihard,Sanakoyeu_2019_CVPR} and face verification \cite{face_verfication_inthewild,sphereface}. %Recent advances in unsupervised representation learning also borrow from DML methodology \cite{moco,pretextmisra}.
Many approaches use ranking-based objectives based on tuples of samples such as pairs \cite{contrastive}, triplets \cite{margin, yu2018correcting}, quadruplets\cite{quadtruplet} or higher-order variants like N-Pairs\cite{npairs}, lifted structure losses \cite{lifted,yu2018correcting} or NCA-based criteria\cite{proxynca}. Further, classification-based methods adjusted to DML \cite{arcface,zhai2018classification} have proven to be effective for learning distance preserving embedding spaces. To address the computational complexity of tuple-based methods\footnote{As an example, the number of triplets scales with $\mathcal{O}(N^3)$, where $N$ is the dataset size.}, different sampling strategies have been introduced \cite{semihard,margin,htl,roth2020pads}. Moreover, proxy-based approaches address this issue by approximating class distributions using only few virtual representatives \cite{proxynca,softriple}. \\
Additionally, more involved research extending above objectives has been proposed: \citet{Sanakoyeu_2019_CVPR} follow a divide-and-conquer strategy by splitting and subsequently merging both the data and embedding space; \citet{abier,dreml} employ an ensemble of specialized learners and \citet{mic,milbich2020diva,milbich2020sharing} combine DML with feature mining or self-supervised learning. Moreover, \citet{dvml} and \citet{hardness-aware} generate artificial samples to effectively augment the training data, thus learning more complex ranking relations. The majority of these methods are trained using the essential objective functions and, further, hinge on the training parameters discussed in our study, thus directly benefiting from our findings. Moreover, we propose an effective regularization technique to improve ranking-based objectives. \\
\textbf{Mini-batch selection:}
The benefits of large mini-batches for training are well studied \citep{smith2017don, goyal2017accurate, keskar2016large}. However, there has been limited research examining effective strategies for the creation of mini-batches.
Research into mini-batch creation has been done to improve convergence in optimization methods for classification tasks\cite{coresetoptim,importance_sampling} or to construct informative mini-batches using core-set selection to optimize generative models \cite{sinha2019small}. Similarly, we analyze mining strategies maximizing data diversity and compare their impact to standard heuristics employed in DML~\cite{margin,mic,Sanakoyeu_2019_CVPR}). \\
\textbf{Generalization in DML:}
Generalization capabilities of representations \cite{alex2016information,shwartzziv2017opening} and, in particular, of discriminative models has been well studied \cite{generalisation_measures,belghazi2018mutual,goyal2017accurate}, e.g. in the light of compression~\cite{tishby2015deep,shwartzziv2017opening} which is covered by strong experimental support~\cite{goyal2019infobot,belghazi2018mutual,alex2016deep}. \citet{manifoldmixup} link compression to a 'flattening' of a representation in the context of classification. We apply this concept to analyze generalization in DML and find that strong compression actually hurts DML generalization. Existing works on generalization in metric learning focus on robustness of linear or kernel-based distance metrics~\cite{Bellet_2015,bellet2013supervised} and examine bounds on the generalization error~\cite{fcn_gen}. In contrast, we examine the correlation between generalization and structural characteristics of the learned embedding space. 
\section{Training a Deep Metric Learning Model}
\label{sec:dml}
In this section, we briefly summarize key components for training a DML model and motivate the main aspects of our study. We first introduce the common categories of training objectives which we consider for comparison in Sec. \ref{sec:objective_function_classes}. Next, in Sec. \ref{sec:batch_sampling} we examine the data sampling process and present strategies for sampling informative mini-batches. Finally, in Sec. \ref{sec:training_params}, we discuss components of a DML model which impact its performance and exhibit an increased divergence in the field, thus impairing objective comparisons.

\subsection{The objective function}
\label{sec:objective_function_classes}
In Deep Metric Learning we learn an embedding function $\phi: \mathcal{X} \mapsto \Phi \subseteq \mathbb{R}^D$ mapping datapoints $x \in \mathcal{X}$ into an embedding space $\Phi$, which allows to measure the similarity between $x_i, x_j$ as $d_{\phi}(x_i,x_j) := d(\phi(x_i), \phi(x_j))$ with $d(.,.)$ being a predefined distance function.
%$\phi(x)$, of datapoints $x \in  \mathbb{R}^{H\times W\times C}$ into an embedding space $\Phi \in \mathbb{R}^D$ which allows to measure the similarity between $x_i, x_j$ as $d_{\phi}(x_i,x_j) := d(\phi(x_i), \phi(x_j))$ with $d$ being a predefined distance function. 
For that, let $\phi := \phi_\theta$ be a deep neural network parametrised by $\theta$ with its output typically normalized to the real hypersphere $\mathbb{S}^D$ for regularization purposes \cite{margin,fcn_gen}. 
In order to train $\phi_\theta$ to reflect the semantic similarity defined by given labels $y \in \mathcal{Y}$, many objective functions have been proposed based on different concepts which we now briefly summarize.\\
\textbf{Ranking-based:} The most popular family are ranking-based loss functions operating on pairs \cite{contrastive}, triplets \cite{semihard, margin} or larger sets of datapoints \cite{npairs, lifted, quadtruplet, rankedlist}. Learning $\phi_\theta$ is defined as an ordering task, such that the distances $d_\phi(x_a, x_p)$ between an anchor $x_a$ and positive $x_p$ of the same class, $y_a = y_p$, is minimized and the distances $d_\phi(x_a, x_n)$ of to negative samples $x_n$ with different class labels, $y_a \neq y_n$, is maximized. For example, triplet-based formulations typically optimize their relative distances as long as a margin $\gamma$ is violated, i.e. as long as $d_\phi(x_a, x_n) - d_\phi(x_a, x_p) < \gamma$. Further, ranking-based objectives are also extended to histogram matching, as proposed in \cite{histogram}. \\
\textbf{Classification-based:}
As DML is essentially solving a discriminative task, some approaches~\cite{zhai2018classification,arcface,sphereface} can be derived from softmax-logits $l_i = W_j^T \phi(x_i) + b_j$. For example, \citet{arcface} exploit the regularization to the real hypersphere $\mathbb{S}^D$ and the equality $W_j^T x_i = \left\lVert W_j^T \right\rVert  \left\lVert \phi(x_i) \right\rVert \cos \varphi_j$ to maximize the margin between classes by direct optimization over angles $\varphi_j$. Further, also standard cross-entropy optimization proves to be effective under normalization \cite{zhai2018classification}.\\
\textbf{Proxy-based:} These methods approximate the distributions for the full class by one~\cite{proxynca} or more~\cite{softriple} learned representatives. By considering the class representatives for computing the training loss, individual samples are directly compared to an entire class. Additionally, proxy-based methods help to alleviate the issue of tuple mining which is encountered in ranking-based loss functions.

\subsection{Data sampling}
\label{sec:batch_sampling}
The synergy between tuple mining strategies and ranking losses has been widely studied~\cite{margin,semihard,htl}. To analyze the impact of data-sampling on performance in the scope of our study, we consider the process of mining informative mini-batches $\mathcal{B}$. This process is independent of the specific training objective and so far has been commonly neglected in DML research. Following we present batch mining strategies operating on both labels and the data itself: \textit{label samplers}, which are sampling heuristics that follow selection rules based on label information only, and \textit{embedded samplers}, which operate on data embeddings themselves to create batches $\mathcal{B}$ of diverse data statistics. \\
\textbf{Label Samplers:} To control the class distribution within $\mathcal{B}$, we examine two different heuristics based on the number, $n$, of 'Samples Per Class' (SPC-$n$) heuristic:\\
\textit{SPC-2/4/8:} Given batch-size $b$, we randomly select $b/n$ unique classes from which we select $n$ samples randomly.\\
\textit{SPC-R:} We randomly select $b-1$ samples from the dataset and choose the last sample to have the same label as one of the other $b-1$ samples to ensure that at least one triplet can be mined from $\mathcal{B}$. Thus, we effectively vary the number of unique classes within mini-batches. \\
%
%\red{with non-trivial extension to a Multi-GPU system due to the common tuple mining process},
\textbf{Embedded Samplers:} 
Increasing the batch-size $b$ has proven to be beneficial for stabilizing optimization due to an effectively larger data diversity and richer training information \cite{coresetoptim,BigGAN,sinha2019small}. As the DML training is commonly performed on a single GPU (limited especially due to tuple mining process on the mini-batch), the batch-size $b$ is bounded by memory. Nevertheless, in order to `virtually' maximize the data diversity, we distill the information content of a large set of samples $\mathcal{B}^*, b^* = |\mathcal{B}^*| > b$ into a mini-batch $B$ by matching the statistics of $\mathcal{B}$ and $\mathcal{B}^*$ under the embedding $\phi$. To avoid computational overhead, we sample $\mathcal{B}^*$ from a continuously updated memory bank $\mathcal{M}$ of embedded training samples. Similar to \citet{pretextmisra}, $\mathcal{M}$ is generated by iteratively updating its elements based on the steady stream of training batches $\mathcal{B}$. Using $\mathcal{M}$, we mine mini-batches by first randomly sampling $\mathcal{B}^*$ from $\mathcal{M}$ with $b^*=1024$ and subsequently find a mini-batch $\mathcal{B}$ to match its data statistics by using one of the following criteria: \\
\textit{Greedy Coreset Distillation (GC):} 
Greedy Coreset~\cite{coresets} finds a batch $\mathcal{B}$ by iteratively adding samples $x^* \in \mathcal{B}^*$ which maximize the distance from the samples that have already been selected $x \in \mathcal{B}$, thereby maximizing the covered space within $\Phi$ by solving  $\min_{\mathcal{B}:|\mathcal{B}|=b} \max_{x*\in \mathcal{B}^*} \min_{x\in \mathcal{B}} d_\phi(x,x^*)$.\\%
% \begin{equation}
% \vspace{-1pt}
%     \min_{\mathcal{B}:|\mathcal{B}|=b} \max_{x*\in \mathcal{B}^*} \min_{x\in \mathcal{B}} d_\phi(x,x^*)
% \end{equation}%
%
\textit{Matching of distance distributions (DDM):} 
DDM aims to preserve the distance distribution of $\mathcal{B}^*$. We randomly select $m$ candidate mini-batches and choose the batch $\mathcal{B}$ with smallest Wasserstein distance between normalized distance histograms of $\mathcal{B}$ and $\mathcal{B}^*$ \cite{emd}.\\
% a normalized distance histogram for the big batch and its $m$ subsampled mini-batchs. We select the mini-batch with the lowest Earth Movers distance to the big batch distance histogramm, see \cite{emd}.\\
\textit{FRD-Score Matching (FRD)}: 
Similar to the recent GAN evaluation setting, we compute the frechet distance~\cite{fid}) between $\mathcal{B}$ and $\mathcal{B}^*$ to measure the similarity between their distributions using $FRD(\mathcal{B},\mathcal{B}^*) = \left\Vert\mu_{\mathcal{B}}-\mu_{\mathcal{B}^*}\right\Vert_2^2 + \text{Tr}(\Sigma_{\mathcal{B}}+\Sigma_{\mathcal{B}^*}-2(\Sigma_{\mathcal{B}}\Sigma_{\mathcal{B}^*})^{1/2})$,
% \begin{align}
%     FRD(\mathcal{B},\mathcal{B}^*) = &\left\Vert\mu_{\mathcal{B}}-\mu_{\mathcal{B}^*}\right\Vert_2^2 + \\&\text{Tr}(\Sigma_{\mathcal{B}}+\Sigma_{\mathcal{B}^*}-2(\Sigma_{\mathcal{B}}\Sigma_{\mathcal{B}^*})^{1/2})
% \end{align}
with $\mu_{\bullet}, \Sigma_{\bullet}$ being the mean and covariance of the embedded set of samples. Like in DDM, we select the closest batch $\mathcal{B}$ to $\mathcal{B}^*$ among $m$ randomly sampled candidates.

%%%%%%%%%%%%%%%%%%%%%%%%%%%%%%%%%%%%%%%%%%%%%%%%%%%%%%%%%%%%%%%%%%%%%%%%%%
\subsection{Training parameters, regularization and architecture}
\label{sec:training_params}

\begin{table}[h]
    \centering
    \begin{tabular}{l|c|c|c}
         Network & GN & IBN & R50\\
         \hline
         CUB200, R@1  & 45.41  & 48.78 & 43.77 \\
         CARS196, R@1 & 35.31  & 43.36 & 36.39 \\
         SOP, R@1     & 44.28  & 49.05 & 48.65 
    \end{tabular}
    \caption{\textit{Recall performance of commonly used network architectures after ImageNet pretraining}. Final linear layer is randomly initialized and normalized.}
    \label{tab:pretrain_perf}
\end{table}

%In this section, we want to address the relevant degree of freedoms in a standard Deep Metric Learning pipeline, which we will also investigate to see how reported results can very when one of these factors is changed:
% While GN and R50 share the same data preprocessing, IBN uses different normalization as well as the BGR format.

Next to the objectives and data sampling process, successful learning hinges on a reasonable choice of the training environment. While there is a multitude of parameters to be set, we identify several factors which both influence performance and exhibit an divergence in lately proposed works. \\
\textit{Architectures}: In recent DML literature predominantly three basis network architectures are used: GoogLeNet \cite{googlenet} (GN, typically with embedding dimensionality 512), Inception-BN \cite{googlenetv2} (IBN, 512) and ResNet50 \cite{resnet} (R50, 128) (with optionally frozen Batch-Normalization layers for improved convergence and stability across varying batch sizes\footnote{Note that Batch-Normalization is still performed, but no parameters are learned.}, see e.g. \citet{mic,learn2rank}). Due to the varying number of parameters and configuration of layers, each architecture exhibits a different starting point for learning, based on its initialization by ImageNet pretraining~\cite{imagenet}. Table \ref{tab:pretrain_perf} compares their initial DML performance measured in Recall@1 (R@1). The reference to differences in architecture is one of the main arguments used by individual works not compare themselves to competing approaches. Disconcertingly, even when reporting additional results using adjusted networks is feasible, typically only results using a single architecture are reported. Consequently, a fair comparison between approaches is heavily impaired. \\
\textit{Weight Decay}: 
Commonly, network optimization is regularized using weight decay/L2-regularization~\cite{weight_decay}. In DML, particularly on small datasets its careful adjustment is crucial to maximize generalization performance. Nevertheless, many works do not report this.\\
\textit{Embedding dimensionality}: 
Choosing a dimensionality $D$ of the embedding space $\Phi$ influences the learned manifold and consequently generalization performance. While each architecture typically uses an individual, standardized dimensionality $D$ in DML, recent works differ without reporting proper baselines using an adjusted dimensionality. Again, comparison to existing works and the assessment of the actual contribution is impaired.\\
\textit{Data Preprocessing}: 
Preprocessing training images typically significantly influences both the learned features and model regularization. Thus, as recent approaches vary in their applied augmentation protocols, results are not necessarily comparable. This includes the trend for increased training and test image sizes.\\
\textit{Batchsize}: 
Batchsize determines the nature of the gradient updates to the network, e.g. datasets with many classes benefit from large batchsizes due to better approximations of the training distribution. However, it is commonly not taken into account as a influential factor of variation.\\
%%%%%%%%%%%%%%%%%%%%%%%%%%%%%%%%%%%%%%%%%%%%%%%%%%%%%%%%%%%%%%%%%%%%%%%%%
\textit{Advanced DML methodologies}
There are many extensions to objective functions, architectures and the training setup discussed so far. However, although extensions are highly individual, they still rely on these components and thus benefit from findings in the following experiments, evaluations and analysis.
%%%%%%%%%%%%%%%%%%%%%%%%%%%%%%%%%%%%%%%%%%%%%%%%%%%%%%%%%%%%%%%%%%%%%%%%%%%

%%%%%%%%%%%%%%%%%%%%%%%%%%%%%%%%%%%%%%%%%%%%%%%%%%%%%%%%%%%%%%%%%%%%%%%%%%%
\section{Analyzing DML training strategies}
\label{sec:eval_ana}

%%%%%%%%%%%%%%%%%%%%%%%%%%%%%%%%%%%%%%%%%%%%%%%%%%%%%%%%%%%%%%%%%%%%%%%%%%
\paragraph{Datasets}
As benchmarking datasets, we use:\\
\textit{CUB200-2011}: Contains 11,788 images in 200 classes of birds. Train/Test sets are made up of the first/last 100 classes (5,864/5,924 images respectively) \cite{cub200-2011}. Samples are distributed evenly across classes.\\ 
\textit{CARS196}: Has 16,185 images/196 car classes with even sample distribution. Train/Test sets use the first/last 98 classes (8054/8131 images) \cite{cars196}.\\ 
\textit{Stanford Online Products (SOP)}: Contains 120,053 product images divided into 22,634 classes. Train/Test sets are provided, contain 11,318 classes/59,551 images in the Train and 11,316 classes/60,502 images in the Test set \cite{lifted}. In SOP, unlike the other benchmarks, most classes have few instances, leading to significantly different data distribution compared to CUB200-2011 and CARS196.

\input{figures/ablations.tex}
\input{figures/meta-eval.tex}
\input{figures/correlation_matrix_metrics}
\subsection{Experimental Protocol}
\label{sec:exp_setup}
Our training protocol follows parts of \citet{margin}, which utilize a ResNet50 architecture with frozen Batch-Normalization layers and embedding dim. 128 to be comparable with already proposed results with this architecture. While both GoogLeNet \cite{googlenet} and Inception-BN \cite{googlenetv2} are also often employed in DML literature, we choose ResNet50 due to its success in recent state-of-the-art approaches~\cite{mic,Sanakoyeu_2019_CVPR}. In line with standard practices we randomly resize and crop images to $224\times 224$ for training and center crop to the same size for evaluation. During training, random horizontal flipping ($p=0.5$) is used. 
Optimization is performed using Adam~\cite{adam} with learning rate fixed to $10^{-5}$ and \textit{no} learning rate scheduling for unbiased comparison.
Weight decay is set to a constant value of $4\cdot 10^{-4}$, as motivated in section \ref{sec:param_study}.
We implemented all models in PyTorch~\cite{pytorch}, and experiments are performed on individual Nvidia Titan X, V100 and T4 GPUs with memory usage limited to 12GB.
Each training is run over 150 epochs for CUB200-2011/CARS196 and 100 epochs for Stanford Online Products, if not stated otherwise.
% Finally, each key result is averaged over five different seeds, while ablation studies are done over three seeds.
For batch sampling we utilize the the SPC-2 strategy, as motivated in section \ref{sec:batch-study}.
Finally, each result is averaged over multiple seeds to avoid seed-based performance fluctuations.
All loss-specific hyperparameters are discussed in the supplementary material, along with their original implementation details.
For our study, we examine the following evaluation metrics (described further in the supplementary): Recall at 1 and 2 \cite{recall}, Normalized Mutual Information (NMI) \cite{nmi}, F1 score \cite{npairs}, mean average precision measured on recall of the number of samples per class (mAP@C) and mean average precision measured on the recall of 1000 samples (mAP@1000). Please see the supplementary (supp. \ref{supp:metrics}) for more information.
%%%%%%%%%%%%%%%%%%%%%%%%%%%%%%%%%%%%%%%%%%%%%%%%%%%%%%%%%%%%%%%%%%%%%%%%%%%
% \subsection{Factors of variation in common DML pipelines}
\subsection{Studying DML parameters and architectures}
\label{sec:param_study}
\input{tables/short_result_table.tex}
Now we study the influence of parameters \& architectures discussed in Sec.~\ref{sec:training_params} using five different objectives. For each experiment, all metrics noted in Sec.~\ref{sec:exp_setup} are measured. For each loss, every metric is normalized by the maximum across the evaluated value range. This enables an aggregated summary of performance across all metrics, where differences correspond to relative improvement.
%For each experiment performance is measured in Recall@1, averaged over 5 runs and normalized by the maximum performance. 
Fig.~\ref{fig:ablation_studies} analyzes each factor by evaluating a range of potential setups with the other parameters fixed to values from Sec.~\ref{sec:exp_setup}: Increasing the batchsize generally improves results with gains varying among criteria, with particularly high relevance on the SOP dataset. For weight decay, we observe loss and dataset dependent behavior up to a relative performance change of $5\%$. Varying the data preprocessing protocol, e.g. augmentations and input image size, leads to large performance differences as well. \textit{Base} follows our protocol described in Sec.~\ref{sec:exp_setup}. \textit{Red.} refers to resizing of the smallest image side to $256$ and cropping to $224$x$224$ with horizontal flipping. \textit{Big} uses \textit{Base} but crops images to $256$x$256$. Finally, we extend \textit{Base} to \textit{Adv.} with color jittering, changes in brightness and hue. We find that larger images provide better performance regardless of objective or dataset. Using the \textit{Adv.} processing on the other hand is dependent on the dataset. Finally, we show that random resized cropping is a generally stronger operation than basic resizing and cropping.\\
All these factors underline the importance of a complete declaration of the training protocol to facilitate reproducibility and comparability. Similar results are observed for the choice of architecture and embedding dimensionality $D$. At the example of R50, our analysis shows that training objectives perform differently for a given $D$ but seem to converge at $D=512$. However, for R50 $D$ is typically fixed to $128$, thus disadvantaging some training objectives over others. Finally, comparing common DML architectures reveals their strong impact on performance with varying variance between loss functions. Highest consistencies seem to be achievable with R50 and IBN-based setups.\\ 
%With their dimensionalities set to $128$ and $512$, respectively, their performance seem interchangable.\\
\textbf{Implications:} In order to warrant unbiased comparability, equal and transparent training protocols and model architectures are essential, as even small deviations can result in large deviations in performance. 
%%%%%%%%%%%%%%%%%%%%%%%%%%%%%%%%%%%%%%%%%%%%%%%%%%%%%%%%%%%%%%%%%%%%%%%%%%%
% The influence of batch-creation for training
\subsection{Batch sampling impacts DML training}
\label{sec:batch-study}
We now analyze how the data sampling process for mini-batches impacts the performance of DML models using the sampling strategies presented in Sec. \ref{sec:batch_sampling}. To conduct an unbiased study, we experiment with six conceptually different objective functions: Marginloss with Distance-Weighted Sampling, Triplet Loss with Random Sampling, ProxyNCA, Multi-Similarity Loss, Histogram loss and Normalized Softmax loss. To aggregate our evaluation metrics (cf. \ref{sec:exp_setup}), we utilize the same normalization procedure discussed in Sec. \ref{sec:param_study}. Fig. \ref{fig:sampling_eval} summarizes the results for each sampling strategy by reporting the distributions of normalized scores of all pairwise combinations of training loss and evaluation metrics. Our analysis reveals that the batch sampling process indeed effects DML training with a difference in \textit{mean} performance up to $1.5\%$. While there is no clear winner across all datasets, we observe that the SPC-2 and FRD samplers perform very well and, in particular, consistently outperform the SPC-4 strategy which is commonly reported to be used in literature \cite{margin,semihard}. \\  
\textbf{Implications:} 
Our study indicates that DML benefits from data diversity in mini-batches, independent of the chosen training objective. This coincides with the general benefit of larger batchsizes as noted in section \ref{sec:param_study}. While complex mining strategies may perform better, simple heuristics like SPC-2 are sufficient. 
%%%%%%%%%%%%%%%%%%%%%%%%%%%%%%%%%%%%%%%%%%%%%%%%%%%%%%%%%%%%%%%%%%%%%%%%%%%
\subsection{Comparing DML models}
Based on our training parameter and batch-sampling evaluations we compare a large selection of $14$ different DML objectives and $4$ mining methods under fixed training conditions (see \ref{sec:exp_setup} \& \ref{sec:param_study}), most of which claim state-of-the-art by a notable margin. For ranking-based models, we employ distance-based tuple mining (D) \cite{margin} which proved most effective. We also include random, semihard sampling \cite{semihard} and a soft version of hard sampling \cite{rothgithub} for our tuple mining study using the classic triplet loss. Loss-specific hyperparameters are determined via small cross-validation gridsearches around originally proposed values to adjust for our training setup. Exact parameters and method details are listed in supp. \ref{supp:criteria}. Table \ref{tab:final_results} summarizes our evaluation results on all benchmarks (\textbf{with other metric rankings} s.a. mAP@C or mAP@1000 \textbf{in the supplementary} (supp. \ref{supp:detailed})), while Fig. \ref{fig:corr_mat} measures correlations between all evaluation metrics. Particularly on CUB200-2011 and CARS196 we find a higher performance saturation between methods as compared to SOP due to strong differences in data distribution. Generally, performance between criteria is much more similar than literature indicates, (see also concurrent work by \citet{musgrave2020metric}). We find that representatives of ranking-based objectives outperform their classification/NCE-based counterparts, though not significantly. On average, margin loss \cite{margin} and multisimilarity loss \cite{multisimilarity} offer the best performance across datasets, though not by a notable margin. Remarkably, under our carefully chosen training setting, a multitude of losses compete or even \textit{outperform} more involved state-of-the-art DML approaches on the SOP dataset. For a detailed comparison to the state-of-the-art, we refer to the supplementary (supp. \ref{supp:sop_sota}). \\
\textbf{Implications:} Under the same setup, performance saturates across methods, contrasting results reported in literature. Taking into account standard deviations, usually left unreported, improvements become even less significant. In addition, carefully trained baseline models are able to outperform state-of-the-art approaches which use considerable stronger architectures. Thus, to evaluate the true benefit of proposed contributions, baseline models need to be competitive and implemented under comparable settings.
%%%%%%%%%%%%%%%%%%%%%%%%%%%%%%%%%%%%%%%%%%%%%%%%%%%%%%%%%%%%%%%%%%%%%%%%%%%
%\subsection{\red{More directions of variance correlate with zero-shot generalization performance}}

\section{Generalization in Deep Metric Learning}
\label{sec:generalization}
The previous section showed how different model and training parameter choices result in vastly different performances. However, how such differences can be explained best on basis of the learned embedding space is an open question and, for instance, studied under the concept of compression~\cite{tishby2015deep}. Recent work \cite{manifoldmixup} links compression to class-conditioned flattening of representation, indicated by an increased decay of singular values obtained by Singular Value Decomposition (SVD) on the data representations. Thus, class representations occupy a more compact volume, thereby reducing the number of directions with significant variance. The subsequent strong focus on the most discriminative directions is shown to be beneficial for classic classification scenarios with i.i.d. train and test distributions. However, this overly discards features which could capture data characteristics outside the training distribution. Hence, generalization in transfer problems like DML is hindered due to the shift in training and testing distribution~\cite{Bellet_2015}. We thus hypothesize that actually retaining a considerable amount of directions of significant variance (DoV) is crucial to learn a well generalizing embedding function $\phi$. \\
%%%%%%
To verify this assumption, we analyze the spectral decay of the embedded training data $\Phi_\mathcal{X} := \{\phi(x)|x\in \mathcal{X}\}$ via SVD. We then normalize the sorted spectrum of singular values (SV) $\mathcal{S}_{\Phi_\mathcal{X}}$\footnote{Excluding highest SV which can obfuscate remaining DoVs.} and compute the KL-divergence to a D-dim. discrete uniform distribution $\mathcal{U}_D$, i.e. $\rho(\Phi) = \text{KL}(\mathcal{U}_D \; || \; \mathcal{S}_{\Phi_\mathcal{X}})$\footnote{For simplicity we use the notation $\rho(\Phi)$ instead of $\rho(\Phi_\mathcal{X})$.}. We don't consider individual training class representations, as testing and training distribution are shifted\footnote{For completeness, class-conditioned singular value spectra as \citet{manifoldmixup} are examined in supp. \ref{supp:mmanifold}.}. Lower values of $\rho(\Phi)$ indicate more directions of significant variance. Using this measure, we analyze a large selection of DML objectives in Fig. \ref{fig:cluster_metric_relations} (rightmost) on CUB200-2011, CARS196 and SOP\footnote{A detailed comparison can be found in supp. \ref{supp:detailed}.}. Comparing R@1 and $\rho(\Phi)$ reveals significant inverse correlation ($\leq -0.63$) between generalization and the spectral decay of embedding spaces $\Phi$, which highlights the benefit of more directions of variance in the presence of train-test distribution shifts.
\\
We now compare our finding to commonly exploited concepts for training such as \textit{(i)} larger margins between classes~\cite{arcface,sphereface}, i.e. an increase in average inter-class distances $\pi_{\text{inter}}(\Phi) = \frac{1}{Z_\text{inter}}\sum_{y_l,y_k,l\neq k} d(\mu(\Phi_{y_l}),\mu(\Phi_{y_k}))$ ; \textit{(ii)} explicitly introducing intra-class variance \cite{dvml}, which is indicated by an increase in average intra-class distance $\pi_{\text{intra}}(\Phi) = \frac{1}{Z_{\text{intra}}}\sum_{y_l\in\mathcal{Y}}\sum_{\phi_i,\phi_j\in \Phi_{y_l},i\neq j} d(\phi_i,\phi_j)$. We also investigate \textit{(iii)} their relation by using the ratio 
$\pi_{\text{ratio}}(\Phi) = \pi_{\text{intra}}(\Phi) / \pi_{\text{inter}}(\Phi)$, which can be regarded as an embedding space density. Here, $\Phi_{y_l} = \{\phi_i := \phi_\theta(x_i) | x_i \in \mathcal{X}, y_i = y_l\}$ denotes the set of embedded samples of a class $y_l$,  $\mu(\Phi_{y_l})$ their mean embedding and $Z_\text{inter}, Z_\text{intra}$ normalization constants. Fig. \ref{fig:cluster_metric_relations} compares these measures with $\rho(\Phi)$. It is evident that neither of the distance related measures $\pi_{\bullet}(\Phi)$ consistently exhibits significant correlation with generalization performance when taking all three datasets into account. For CUB200-2011 and CARS196, we however find that an increased embedding space density ($\pi_\text{ratio})$ is linked to stronger generalisation. For SOP, its estimate is likely too noisy due to the strong imbalance between dataset size and amount of samples per class.\\ 
\textbf{Implications:} Generalization in DML exhibits strong inverse correlation to the SV spectrum decay of learned representations, as well as a weaker correlation to the embedding space density. This indicates that representation learning under considerable shifts between training and testing distribution is hurt by excessive feature compression, but may benefit from a more densely populated embedding space. 

%We do note however that with increasing dataset size, especially on the large SOP dataset, $\pi_{intra}$ and $\pi_{inter}$ exhibit notable correlation to generalization performance. We attribute this to high sample diversity, which results in higher embedding space coverage (as noted by the near constant density ratio $\pi_{ratio}$). This property has been leveraged for successful face-retrieval methods \cite{arcface,sphereface,cosface}, where evaluation benchmarks exhibit similar class distributions like SOP.

\subsection{$\rho$-regularization for improved generalization}
\input{figures/cluster_metric_relations.tex}
\input{figures/toy_ex}
\input{figures/singular_vec_progression.tex}
We now exploit our findings to propose a simple $\rho$-regularization for ranking-based approaches by counteracting the compression of representations. We randomly alter tuples by switching negative samples $x_n$ with the positive $x_p$ in a given ranking-loss formulation (cf. Sec. \ref{sec:objective_function_classes}) with probability $p_{\text{switch}}$. %For triplets, $x_p$ is replaced with anchor $x_a$ to allow for compatibility with SPC-2 sampling, ensuring that $x_p\neq x_n$ (effectively converting it to a contrastive loss between two samples of the same class). 
This pushes samples of the same class apart, enabling a DML model to capture extra non-label-discriminative features while dampening the compression induced by strong discriminative training signals.\\ 
Fig. \ref{fig:toy_example} depicts a 2D toy example (details supp. \ref{supp:more_toy}) illustrating the effect of our proposed regularization while highlighting the issue of overly compressed data representations. Even though the training distribution exhibits features needed to separate all test classes, these features are disregarded by the strong discriminative training signal. Regularizing the compression by attenuating the spectral decay $\rho(\Phi)$ enables the model to capture more information and exhibit stronger generalization to the unseen test classes. In addition, Fig. \ref{fig:singular_value_progression} verifies that the $\rho$-regularization also leads to a decreased spectral decay on DML benchmark datasets, resulting in improved recall performance (cf. Tab. \ref{tab:final_results} (bottom)), while being reasonably robust to changes in $p_\text{switch}$ (see supp. \ref{supp:switch_ablate}). In contrast, in the appendix we also see that encouraging higher compression seems to be detrimental to performance.\\
\textbf{Implications:} Implicitly regularizing the number of directions of significant variance can improve generalization.

%%%%%%%%%%%%%%%%%%%%%%%%%%%%%%%%%%%%%%%%%%%%%%%%%%%%%%%%%%%%%%%%%%%%%%%%%%
%%%%%%%%%%%%%%%%%%%%%%%%%%%%%%%%%%%%%%%%%%%%%%%%%%%%%%%%%%%%%%%%%%%%%%%%%%%
\section{Conclusion}
\label{sec:conclusion}
In this work, we counteract the worrying trend of diverging training protocols in Deep Metric Learning (DML). 
% We conduct a large, comprehensive study of the most influential components for training a DML model which, further, contribute severely to the impaired comparability of recent approaches. 
We conduct a large, comprehensive study of important training components and objectives in DML to contribute to improved comparability of recent and future approaches. 
% Next to our study, 
On this basis,
% we analyze the correlation between DML generalization and the compression of the learned data representation. 
we study generalization in DML and uncover a strong correlation to the level of compression and embedding density of learned data representation. 
% Our findings uncover that strongly discriminative training signals disregard crucial features for capturing data characteristics that transfer to unknown distributions. 
Our findings reveal that highly compressed representations disregard helpful features for capturing data characteristics that transfer to unknown test distributions.
To this end, we propose a simple technique for ranking-based methods to regularize the compression of the learned embedding space, which results in boosted performance across all benchmark datasets.

\section*{Acknowledgements}
We would like to thank Alex Lamb for insightful discussions, and Sharan Vaswani \& Dmitry Serdyuk for their valuable feedback (all Mila). We also thank Nvidia for donating NVIDIA DGX-1, and Compute Canada for providing resources for this research.

\section*{Reviewer Comments}
\textbf{Re: Data Augmentation/Different Spatial Resolution:}
We agree that data augmentation is an important factor to regularize training. We have made sure to analyze the impact of different image augmentations and resolutions on the performance.

\textbf{Re: In-Shop dataset experiments:}
The data distribution of the In-Shop dataset is very similar to the SOP dataset, thus relative results are generally transferable, which is why we have decided not to include it in this work.

\textbf{Re: Grammar and typos:}
We thank the reviewers for pointing out these errors and have made sure to correct them.

\textbf{Re: Missing references:}
All mentioned references have been added.

\bibliography{references}
\bibliographystyle{icml2020}

\appendix
\input{supplementary_arxiv}

\end{document}

%% file: figures/first_page.tex
%\begin{figure}[t]
%\centering
%\begin{subfigure}[c]{1\linewidth}
%\begin{center}
%\includegraphics[width=0.8\linewidth]{images/first_page_top.pdf}
%\end{center}
%   \caption{Comparison of Existing Criteria and Influence of our proposed density regularization.}
%\label{fig:fp1}
%\end{subfigure}
%\begin{subfigure}[c]{1\linewidth}
%\begin{center}
%\includegraphics[width=0.8\linewidth]{images/first_page_bottom.pdf}
%\end{center}
%   \caption{Evaluation of batch creation criteria.}
%\label{fig:first_page_batch_sampling}
%\end{subfigure}
%\end{figure}

\begin{figure}[t]
\centering
%\begin{subfigure}[c]{1\linewidth}
\begin{center}
\includegraphics[width=0.99\linewidth]{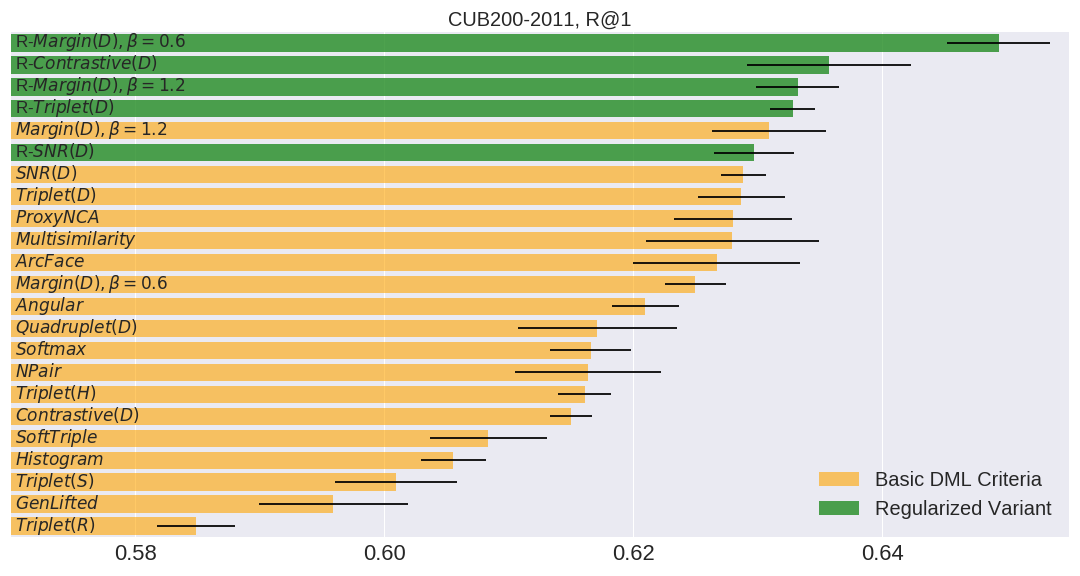}
\end{center}
\vspace{-5pt}
   \caption{\textit{Mean recall performance and standard deviation} of various DML objectives trained with (green) and without (orange) our proposed regularization. For all benchmarks, see appendix.}
\label{fig:fp1}
%\end{subfigure}
\vspace{-12pt}
\end{figure}

%% file: figures/ablations.tex
\begin{figure}[t]
\centering
\begin{subfigure}[c]{1\linewidth}
\begin{center}
\includegraphics[width=1\linewidth]{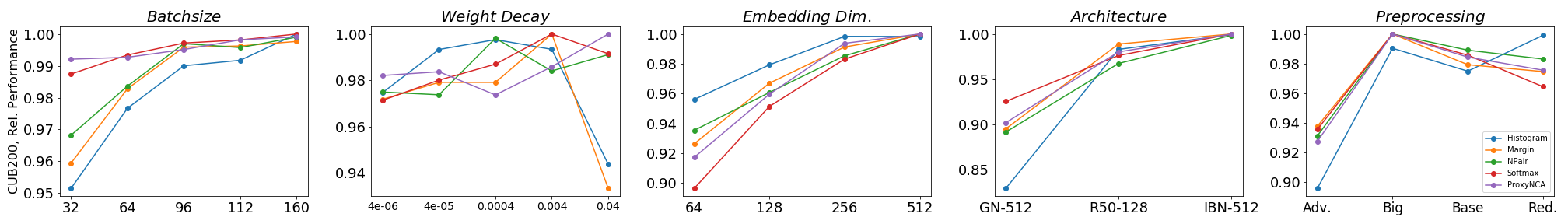}
\end{center}
\end{subfigure}
\begin{subfigure}[c]{1\linewidth}
\begin{center}
\includegraphics[width=1\linewidth]{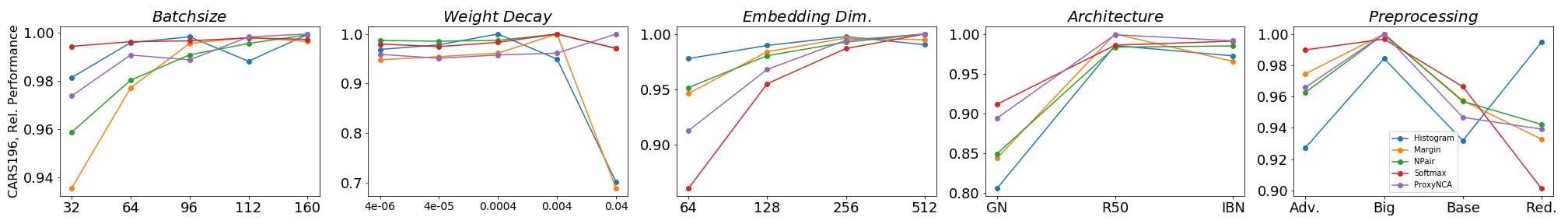}
\end{center}
\end{subfigure}
\begin{subfigure}[c]{1\linewidth}
\begin{center}
\includegraphics[width=1\linewidth]{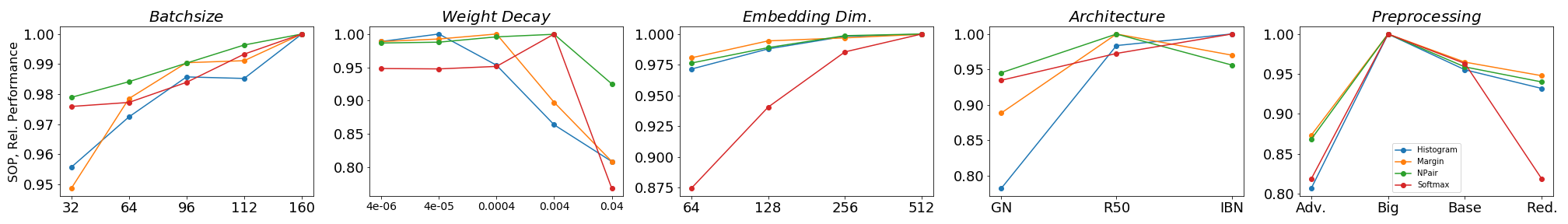}
\end{center}
\end{subfigure}
\caption{\textit{Evaluation of DML pipeline parameters and architectures} on all benchmark datasets and their influence on relative improvement across different training criteria.}
\label{fig:ablation_studies}
\end{figure}

%% file: figures/meta-eval.tex
\begin{figure*}[t]
\centering
\begin{subfigure}[c]{0.33\linewidth}
\centering
\includegraphics[width=1\textwidth]{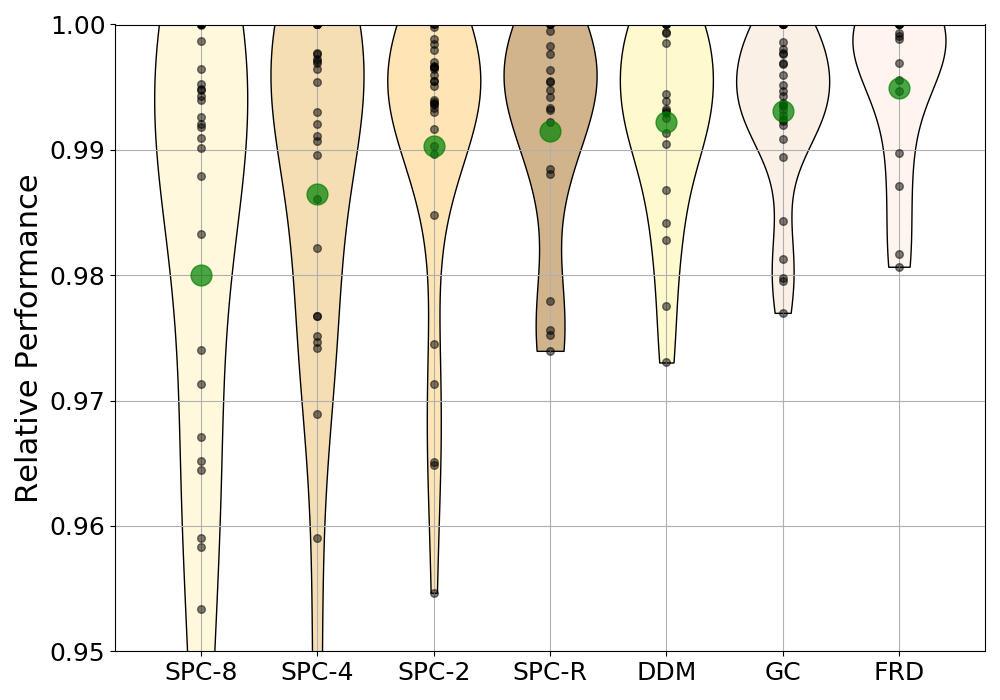}
\caption{CUB200-2011}
\end{subfigure}
\begin{subfigure}[c]{0.33\linewidth}
\centering
\includegraphics[width=1\textwidth]{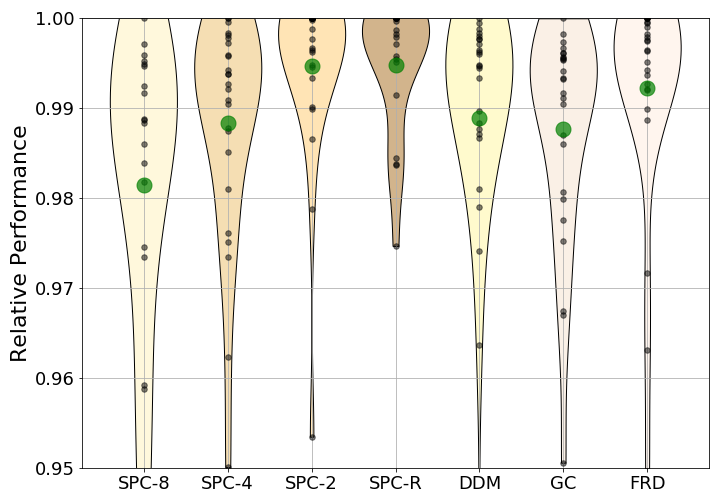}
\caption{CARS196}
\end{subfigure}
\begin{subfigure}[c]{0.33\linewidth}
\centering
\includegraphics[width=1\textwidth]{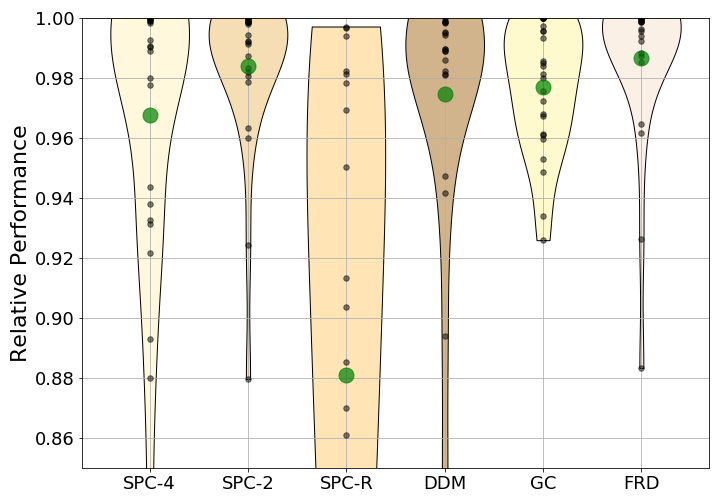}
\caption{SOP}
\end{subfigure}
\caption{\textit{Comparison of mini-batch mining strategies} on three different datasets. Performance measures Recall@1 and 2, NMI, mAP and F1 are normalized across metrics and loss function. We plot the distributions of relative performances for each strategy.}
\label{fig:sampling_eval}
\vspace{-6pt}
\end{figure*}

% On seven different criteria, we evaluate the performance change across Recall@1 and 2, NMI, mAP and F1. For each criterion and each metric, the scores are normalized over the max-value. As an example, assume ProxyNCA and NMI - for this setup, have a score for each batch-creation methods. These scores are then normalized to the maximum value. Doing this for each loss and metrics allows to do an aggregate comparison over losses and metrics for each method.

%% file: figures/correlation_matrix_metrics.tex
\begin{figure}[h]
\begin{center}
\includegraphics[width=1\linewidth]{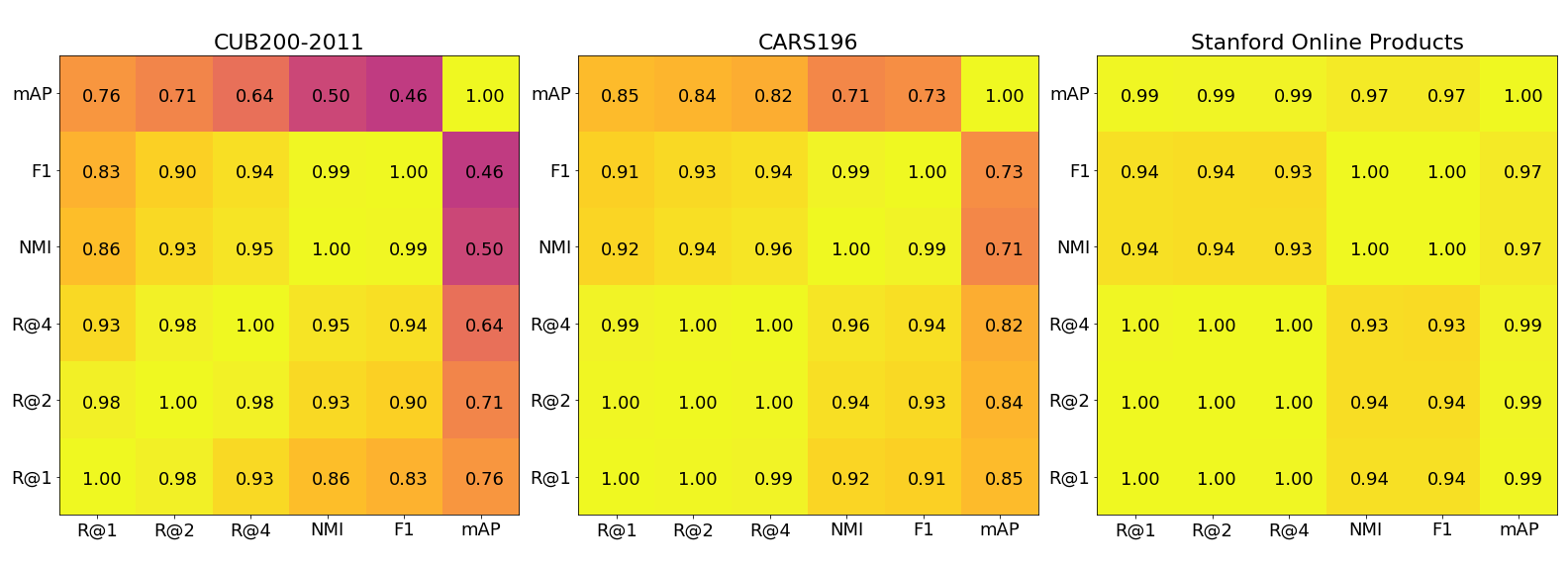}
\end{center}
\vspace{-6pt}
   \caption{\textit{Metrics Correlation matrix} for standard (Recall, NMI) and underreported retrieval metrics. mAP denotes mAP@C. Please refer to the supplementary for more information.}
\label{fig:corr_mat}
\vspace{-10pt}
\end{figure}

%% file: tables/short_result_table.tex
\begin{table*}[t]
 \small
 \rowcolors{2}{gray!8}{white}
   \setlength\tabcolsep{1.4pt}
   \centering
   %\begin{tabular}{l|c|ccc|c}
   \begin{tabular}{l|c|c||c|c||c|c}
     \toprule
     \multicolumn{1}{l}{Benchmarks$\rightarrow$} & \multicolumn{2}{c}{CUB200-2011} & \multicolumn{2}{c}{CARS196} & \multicolumn{2}{c}{SOP} \\
     \midrule
     Approaches $\downarrow$ & R@1 & NMI & R@1 & NMI & R@1 & NMI\\
     \midrule
    Imagenet~\cite{imagenet} & $43.77$ & $57.56$ & $36.39$ & $37.96$ & $48.65$ & $58.64$ \\
    \midrule
    Angular~\cite{angular} & $62.10\pm0.27$ & $67.59\pm0.26$ & $78.00\pm0.32$ & $66.48\pm0.44$ & $73.22\pm0.07$ & $89.53\pm0.01$\\
    ArcFace~\cite{arcface} & $62.67\pm0.67$ & $67.66\pm0.38$ & $79.16\pm0.97$ & $66.99\pm1.08$ & $77.71\pm0.15$ & $90.09\pm0.03$\\
    Contrastive~\cite{contrastive} (D) & $61.50\pm0.17$ & $66.45\pm0.27$ & $75.78\pm0.39$ & $64.04\pm0.13$ & $73.21\pm0.04$ & $89.78\pm0.02$\\
    GenLifted~\cite{genlifted} & $59.59\pm0.60$ & $65.63\pm0.14$ & $72.17\pm0.38$ & $63.75\pm0.35$ & $75.21\pm0.12$ & $89.84\pm0.01$\\
    Hist.~\cite{histogram} & $60.55\pm0.26$ & $65.26\pm0.23$ & $76.47\pm0.38$ & $64.15\pm0.36$ & $71.30\pm0.10$ & $88.93\pm0.02$\\
    Margin (D, $\beta=0.6$)~\cite{margin} & $62.50\pm0.24$ & $67.02\pm0.37$ & $77.70\pm0.32$ & $65.29\pm0.32$ & $77.38\pm0.11$ & \blue{$\mathbf{90.45\pm0.03}$}\\
    Margin (D, $\beta=1.2$)~\cite{margin} & $\mathbf{63.09\pm0.46}$ & $68.21\pm0.33$ & $79.86\pm0.33$ & $67.36\pm0.34$ & $\mathbf{78.43\pm0.07}$ & $90.40\pm0.03$\\
    Multisimilarity~\cite{multisimilarity} & $62.80\pm0.70$ & \blue{$\mathbf{68.55\pm0.38}$} & $\mathbf{81.68\pm0.19}$ & \blue{$\mathbf{69.43\pm0.38}$} & $77.99\pm0.09$ & $90.00\pm0.02$\\
    Npair~\cite{npairs} & $61.63\pm0.58$ & $67.64\pm0.37$ & $77.48\pm0.28$ & $66.55\pm0.19$ & $75.86\pm0.08$ & $89.79\pm0.03$\\
    Pnca~\cite{proxynca} & $62.80\pm0.48$ & $66.93\pm0.38$ & $78.48\pm0.58$ & $65.76\pm0.22$ & $-$ & $-$\\
    Quadruplet (D)~\cite{quadtruplet} & $61.71\pm0.63$ & $66.60\pm0.41$ & $76.34\pm0.27$ & $64.79\pm0.50$ & $76.95\pm0.10$ & $90.14\pm0.02$\\
    SNR (D)~\cite{signal2noise} & $62.88\pm0.18$ & $67.16\pm0.25$ & $78.69\pm0.19$ & $65.84\pm0.52$ & $77.61\pm0.34$ & $90.10\pm0.08$\\
    SoftTriple~\cite{softriple} & $60.83\pm0.47$ & $64.27\pm0.36$ & $75.66\pm0.46$ & $62.66\pm0.16$ & $-$ & $-$\\
    Softmax~\cite{zhai2018classification} & $61.66\pm0.33$ & $66.77\pm0.36$ & $78.91\pm0.27$ & $66.35\pm0.30$ & $76.92\pm0.64$ & $89.82\pm0.15$\\
    \midrule
    Triplet (D)~\cite{margin} & $62.87\pm0.35$ & $67.53\pm0.14$ & $79.13\pm0.27$ & $65.90\pm0.18$ & $77.39\pm0.15$ & $90.06\pm0.02$\\
    Triplet (H)~\cite{rothgithub} & $61.61\pm0.21$ & $65.98\pm0.41$ & $77.60\pm0.33$ & $65.37\pm0.26$ & $73.50\pm0.09$ & $89.25\pm0.03$\\
    Triplet (R)~\cite{semihard} & $58.48\pm0.31$ & $63.84\pm0.30$ & $70.63\pm0.43$ & $61.09\pm0.27$ & $67.86\pm0.14$ & $88.35\pm0.04$\\
    Triplet (S)~\cite{semihard} & $60.09\pm0.49$ & $65.59\pm0.29$ & $72.51\pm0.47$ & $62.84\pm0.41$ & $73.61\pm0.14$ & $89.35\pm0.02$\\
    \midrule
    \midrule
    R-Contrastive (D) & $63.57\pm0.66$ & $67.63\pm0.31$ & $81.06\pm0.41$ & $67.27\pm0.46$ & $74.36\pm0.11$ & $89.94\pm0.02$\\
    R-Margin (D, $\beta=0.6$) & \blue{$\mathbf{64.93\pm0.42}$} & $68.36\pm0.32$ & \blue{$\mathbf{82.37\pm0.13}$} & $68.66\pm0.47$ & $77.58\pm0.11$ & $90.42\pm0.03$\\
    R-Margin (D, $\beta=1.2$) & $63.32\pm0.33$ & $67.91\pm0.66$ & $81.11\pm0.49$ & $67.72\pm0.79$ & \blue{$\mathbf{78.52\pm0.10}$} & $90.33\pm0.02$\\
    R-SNR (D) & $62.97\pm0.32$ & $68.04\pm0.34$ & $80.38\pm0.35$ & $67.60\pm0.20$ & $77.69\pm0.25$ & $90.02\pm0.06$\\
    R-Triplet (D) & $63.28\pm0.18$ & $67.86\pm0.51$ & $81.17\pm0.11$ & $67.79\pm0.23$ & $77.33\pm0.14$ & $89.98\pm0.04$\\
    \bottomrule
    \end{tabular}
    \caption{\textit{Comparison of Recall@1 and NMI performances for all objectives} averaged over 5 runs. Each model is trained using the same training setting over 150 epochs for CUB/CARS and 100 epochs for SOP. 'R-' denotes model trained with $\rho$-regularization. \textbf{Bold} denotes best results excluding regularization. \blue{\textbf{Boldblue}} marks overall best results. \textbf{Please note that a ranking on all other metrics (s.a. mAP@C, mAP@1000) as well as a visual summary can be found in the supplementary} (supp. \ref{supp:detailed}, supp. \ref{supp:visualsummary})!}
    \label{tab:final_results}
\vspace{-8pt}
 \end{table*}

%% file: figures/cluster_metric_relations.tex
\begin{figure}[t]
\begin{center}
\includegraphics[width=1\linewidth]{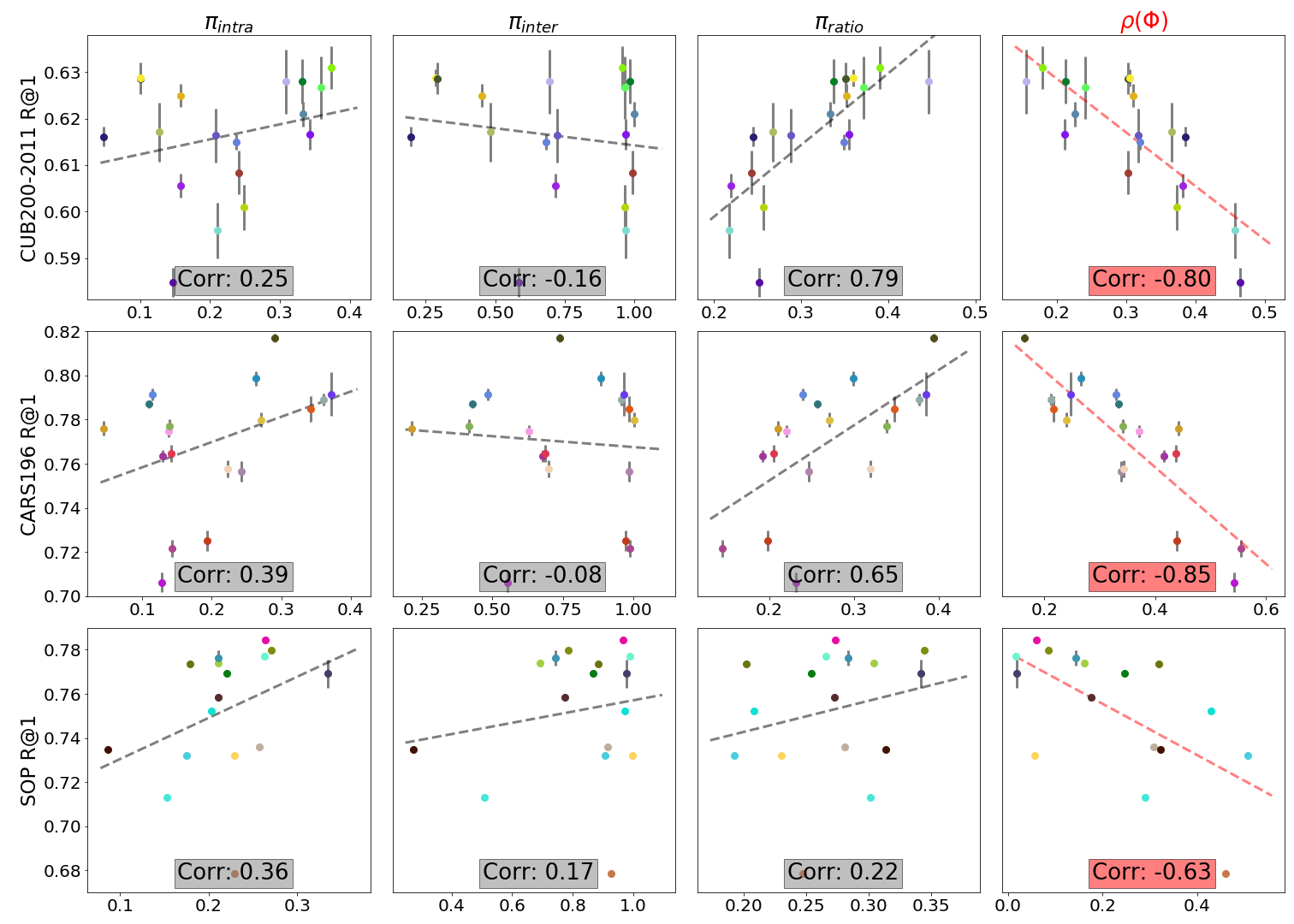}
\end{center}
\vspace{-5pt}
   \caption{\textit{Correlation between generalization and structural properties} derived from $\Phi_{\mathcal{X}}$ using different DML objectives on each dataset. \textit{Left-to-Right}: Mean intra-class distances $\pi_{\text{intra}}$ \& inter-class distances $\pi_{\text{inter}}$, the ratio $\pi_{\text{intra}} / \pi_{\text{inter}}$ and spectral decay $\rho$. }
\label{fig:cluster_metric_relations}
\vspace{-5pt}
\end{figure}
% Performance relation between various embedding space metrics and generalization performance across DML criteria. Each colored point corresponds to a specific criterion. Comparison by color is possible within a row, but not completely inbetween.

%% file: figures/toy_ex.tex
\begin{figure}[t]
\begin{center}
\includegraphics[width=1\linewidth]{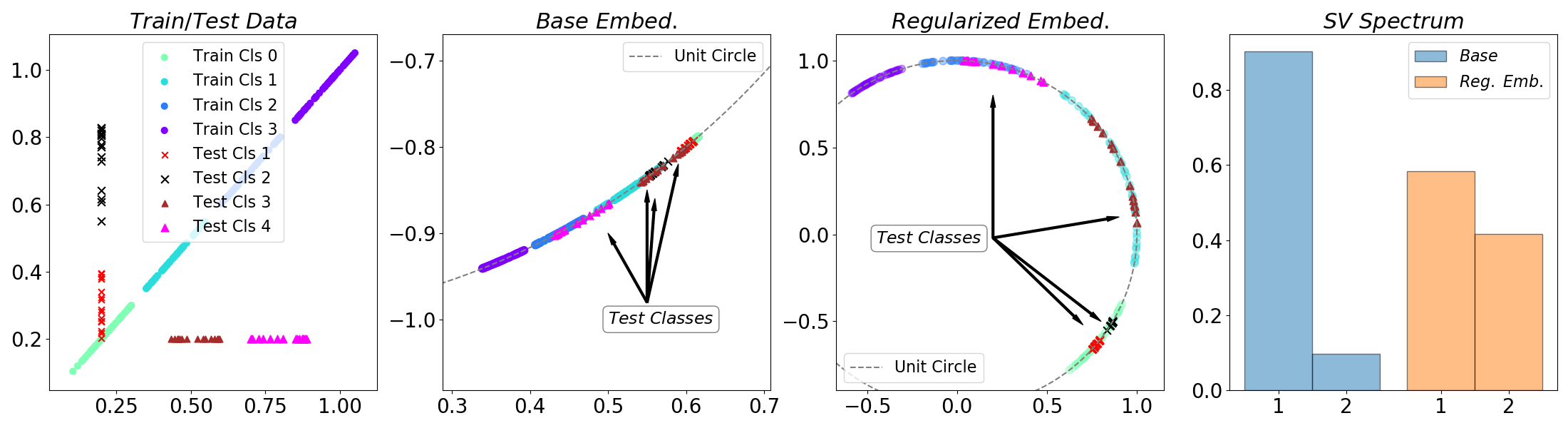}
\end{center}
   \caption{\textit{Toy example illustrating the effect of} $\rho$-regularization. (Leftmost) training and test data. (Mid-left) A small, normalized two-layer fully-connected network trained with standard contrastive loss fails to separate all test classes due to excessive compression of the learned embedding. (Mid-right) The regularized embedding successfully separates the test classes by introducing a lower spectral decay. (Rightmost) Singular value spectra of training embeddings learned with and without regularization.}
\label{fig:toy_example}
\end{figure}

%% file: figures/singular_vec_progression.tex
\begin{figure}[t]
\begin{center}
\includegraphics[width=1\linewidth]{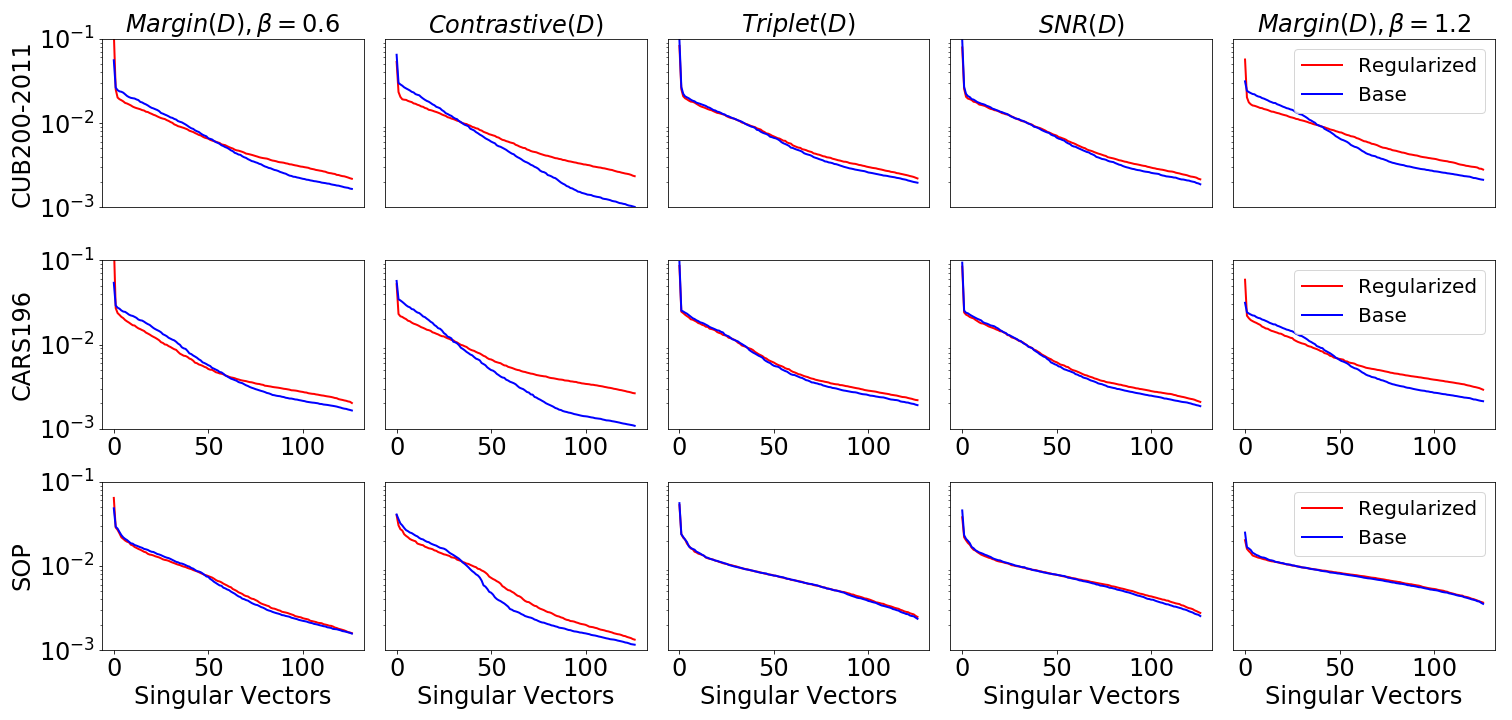}
\end{center}
\vspace{-8pt}
   \caption{\textit{Sing. Value Spectrum} for models trained with (red) and without (blue) $\rho$-regularization for various ranking-based criteria.}
\label{fig:singular_value_progression}
\vspace{-8pt}
\end{figure}

% Change in SVD Spectrum of regularized and non-regularized criteria. It can be seen that generalization improves with more introduced directions of variance.

%% file: supplementary_arxiv.tex
\onecolumn
\icmltitle{Supplementary: Revisiting Training Strategies and Generalization Performance in Deep Metric Learning}

%%%%%%%%%%%%%%%%%%%%%%%%%%%%%%%%%%%%%%%%%%%%%%%%%%%%%%%%%%%
\section{Description of Methods}
In this section, we briefly describe each DML training objective and triplet mining strategy used in our study, as well as the choice of their individual hyperparameters. General training parameters and details of the training protocol are already discussed in the main paper in Sec. 4.1. For notation, we refer to the embedding of an image $x_i$ including output normalization as $\phi_i=\phi(x_i)$. The non-normalized version is denoted as $\phi^*_i$. All methods operate on the mini-batch $\mathcal{B}$ containing image indices. If not mentioned otherwise, all embeddings operate in dimension $D=128$.
%%%
\subsection{Training Criteria}\label{supp:criteria}
\paragraph{Contrastive~\cite{contrastive}} The contrastive training formalism is simple: Given embedding pairs $\mathcal{P}$ (sampled from a mini-batch of size $b$) containing an anchor $\phi_a$ from class $y_a$ and either a positive $\phi_p$ with $y_p=y_a$ or a negative $\phi_n$ from a different class, $y_n\neq y_a$, the network $\phi$ is trained to minimize
\begin{equation}\label{eq:contrastive}
    \mathcal{L}_{contr} = \frac{1}{b}\sum^b_{(i,j) \in \mathcal{P}} \mathbb{I}_{y_i=y_j} d_e(\phi_i, \phi_j) + \mathbb{I}_{y_i\neq y_j} [\gamma - d_e(\phi_i, \phi_j)]_+
\end{equation}
with margin $\gamma$, which we set to $1$. The margin ensures that embeddings are not projected arbitrarily far apart from each other. For our distance function we utilize the standard euclidean distance $d_e(x,y) = \left\Vert x-y \right\Vert_2$. We combine the contrastive loss with the distance-weighting negative sampling mentioned below.
%%%
\paragraph{Triplet~\cite{face_verfication_inthewild}} Triplets extend the contrastive formalism to provide a concurrent ranking surrogate for both negative and positive sample embeddings using triplets $\mathcal{T}$ sampled from a mini-batch:
\begin{equation}
    \mathcal{L}_{tripl} = \frac{1}{b} \sum^{b}_{\substack{(a,p,n)\in\mathcal{T} \\ y_a=y_p\neq y_n}} \left[d_e(\phi_a, \phi_p) - d_e(\phi_a, \phi_n) + \gamma\right]_+
\end{equation}
with margin $\gamma=0.2$, thus following recent implementations in e.g. \citet{mic} or \cite{margin}.  Initial works \cite{semihard} using the triplet loss commonly utilized random or semihard triplet sampling and a GoogLeNet-based architecture. Recent methods typically employ the more effective distance-weighted sampling~\cite{margin} and more powerful networks~\cite{mic,Sanakoyeu_2019_CVPR}. For completeness, we compare the triplet-loss performance combinde with random, semihard and distance-weighted sampling schemes introduced below.
%%%
\paragraph{Generalized Lifted Structure~\cite{genlifted}}
The Generalized Lifted Structure loss extends the standard lifted structure loss \cite{lifted} to include all available anchor-positive and anchor-negative distance pairs within a mini-batch $\mathcal{B}$, instead of utilizing only a single anchor-positive combination:
\begin{equation}
    \mathcal{L}_{genlift} = \sum_{a\in\mathcal{B}} \left[ \log \sum_{p \in \mathcal{B}, y_a=y_p} \exp{\left(d_e(\phi^*_a, \phi^*_p)\right)} + \log \sum_{n \in \mathcal{B}, y_n \neq y_a} \exp{\left(\gamma-d_e(\phi^*_a,\phi^*_n)\right)}\right]_+ + \frac{\nu}{b}\cdot \sum_{a\in \mathcal{B}} \left\Vert \phi^*_a \right\Vert_2^2 
\end{equation}

with mini-batch samples of a class $c$ grouped into $C$, and sets of $C$ contained in $\mathcal{C}$. $\phi^*$ denotes the non-normalized version of $\phi$. The margin $\gamma=1$ serves the standard purpose of avoiding over-distancing already correct image pairs. To account for increasing values, $\nu=0.005$ regularizes the embeddings. 

\paragraph{N-Pair~\cite{npairs}} N-Pair or N-Tuple losses extend the triplet formalism to incorporate all negatives in the mini-batch $\mathcal{B}$ by
\begin{equation}
    \mathcal{L}_{npair} = \frac{1}{b} \sum_{\substack{(a,p) \in \mathcal{B} \\ y_a=y_p, a\neq p}} \log \left(1+\sum_{\substack{n \in \mathcal{B} \\ y_a\neq y_n}} \exp{(\phi_a^{*,T}\phi_n - \phi_a^{*,T}\phi^*_p)}\right) + \frac{\nu}{b}\cdot \sum_{i\in \mathcal{B}} \left\Vert \phi^*_i \right\Vert_2^2 
\end{equation}
with embedding regularization $\nu=0.005$, as \cite{npairs} noted a slow convergence for normalized embeddings.
%%%
\paragraph{Angular~\cite{angular}}   
By introducing an angle-based penalty, the angular loss effectively introduces scale invariance and higher-order geometric constraints that are not explicitly introduced in normal contrastive losses:
\begin{equation}
    \mathcal{L}_{ang} = \mathcal{L}_{npair}(\phi^*) + \frac{\lambda}{b} \sum_{\substack{(a,p)\in\mathcal{B}\\y_a=y_p, a\neq p}}\left[\log (1+\sum_{\substack{n\in\mathcal{B}\\y_n\neq y_a}} \exp{\left(4\tan^2\left(\alpha(\phi_a + \phi_p)^T\phi_n\right)\right) - 2\left(1+\tan^2(\alpha)\right)\phi_a^T\phi_n} \right] 
\end{equation}
with angular margin $\alpha$, which, as proposed in the original paper, is set to $\pi/4$. $\lambda=2$ is the trade-off between standard ranking losses and the angular constraint. The N-Pair parameters are set as above.
%%%
\paragraph{Arcface~\cite{arcface}}
Arcface transforms the standard softmax formulation typically used in classification problem to retrieval-based problems by enforcing an angular margin between the embeddings $\phi$ and an approximate center $W\in\mathbb{R}^{c\times d}$ for each class, resulting in 
\begin{equation}
    \mathcal{L}_{arc} = -\frac{1}{b}\sum_{i\in\mathcal{B}}\log \frac{\exp(s\cdot\cos{(W_{y_i}^T\phi_i+\gamma=0.5)})}{\exp{(s\cdot\cos{(W_{y_i}^T\phi_i+\gamma=0.5)})} + \sum_{\substack{j\in\mathcal{B}\\y_i\neq y_j}} \exp{(s\cdot\cos{(W_{y_j}^T\phi_i)})}}
\end{equation}
Further, this training objective also introduces the additive angular margin penalty $\gamma=0.5$ for increased inter-class discrepancy, while the scaling $s=16$ denotes the radius of the effective utilized hypersphere $\mathbb{S}$. The class centers are optimized with learning rate $0.0005$.
%%%
\paragraph{Histogram~\cite{histogram}}
In contrast to many sample-based ranking objective functions, Histogram Loss learns to minimize the probability of a positive sample pair having a higher similarity score than a negative pair. Given a mini-batch $\mathcal{B}$, the sets of positive similarities $\mathcal{S}^+ = \{\phi_i^T\phi_j | (i,j)\in\mathcal{P}, y_i=y_j\}$ and negative similarities $\mathcal{S}^- = \{\phi_i^T\phi_j | (i,j)\in\mathcal{P},  y_i\neq y_j\}$, one optimises

\begin{align}
    \delta(s,r) &= \frac{1}{\Delta} \left(\mathbb{I}_{s\in[t_{r-1},t_r]}\cdot(s-t_{r-1}) + \mathbb{I}_{s\in[t_{r-1},t_r]}\cdot(t_{r+1}-s)\right)\\
    h^{+/-}(r)    &= \frac{1}{\Vert \mathcal{S}^{+/-} \Vert} \sum_{s\in \mathcal{S}^{+/-}} \delta(s, r)\\
    \mathcal{L}_{hist} &= \sum_{r\in R} h^-(r) \left( \sum_{q=1}^r h^+(q) \right)
\end{align}

resulting in soft, differentiable histogram assignments. The final objective $\mathcal{L}_{hist}$ then penalizes strong overlap between the probability of positive pairs having higher distance (i.e. its cumulative distribution to point $r$) than respective negative pairs.
Such a histogram loss introduces a single hyperparameter, namely the degree of histogram discretisation $R$, which we set to $65$ for CUB200-2011 and CARS196 and $11$ for SOP in our study. In general, our implementation borrows from the original code base used in \cite{histogram}. 

\paragraph{Margin~\cite{margin}}
Margin loss extends the standard triplet loss by introducing a dynamic, learnable boundary $\beta$ between positive and negative pairs. This transfers the common triplet ranking problem to a relative ordering of pairs $\mathcal{P} = \{(i,j) | i,j \in \mathcal{B}, y_i \neq y_j\}$:
\begin{equation}
    \mathcal{L}_{margin} = \sum_{(i,j) \in \mathcal{P}} \gamma + \mathbb{I}_{y_i=y_j}(d(\phi_i, \phi_j) - \beta) - \mathbb{I}_{y_i\neq y_j}(d_e(\phi_i, \phi_j)-\beta) 
\end{equation}
The learning rate of the boundary $\beta$ is set to $0.0005$, with initial value either $0.6$ or $1.2$ and triplet margin $\gamma=0.2$. For our implementation, we utilise the distance-weighted triplet sampling method highlighted below.
%%%
\paragraph{MultiSimilarity~\cite{multisimilarity}}
Unlike contrastive and triplet based ranking methods, the MultiSimilarity loss concurrently evaluates similarities between anchor and negative, anchor and positive, as well as positive-positive and negative-negative pairs in relation to an anchor:
\begin{align}
    s^*_c(i,j) &= \begin{cases}
                    s_c(\phi_i,\phi_j) \qquad s_c(\phi_i,\phi_j) > \min_{j\in\mathcal{P}_i}s_c(\phi_i,\phi_j) - \epsilon\\
                    s_c(\phi_i,\phi_j) \qquad s_c(\phi_i,\phi_j) < \max_{k\in\mathcal{N}_i}s_c(\phi_i,\phi_k) + \epsilon\\
                    0\qquad\qquad\qquad\text{otherwise}
                    \end{cases}\\
    \mathcal{L}_{multisim} &= \frac{1}{b} \sum_{i\in \mathcal{B}} \left[ \frac{1}{\alpha}\log[1+\sum_{j\in\mathcal{P}_i}\exp(-\alpha(s^*_c(\phi_i,\phi_j)-\lambda))] + \frac{1}{\beta}\log[1+\sum_{k\in\mathcal{N}_i}\exp(\beta(s^*_c(\phi_i,\phi_k)-\lambda))] \right]
\end{align}
where $\mathcal{P}_x$ and $\mathcal{N}_x$ denote the set of positive and negative samples for a sample $x$, with cosine similarity $s_c(x,y) = x^Ty$ for two normalized vectors $x,y\in\mathbb{R}^d$. For our hyperparameters, we use $\alpha=2$, $\beta=40$, $\lambda=0.5$ and $\epsilon=0.1$.
%%%
\paragraph{ProxyNCA~\cite{proxynca}}
The sampling complexity of tuples heavily affects the training convergence. ProxyNCA introduces a remedy by introducing class proxies, which act as approximations to entire classes. This way only an anchor is sampled and compared against the respective positive and negative class proxies. Utilizing one proxy $\psi_c\in\mathbb{R}^d$ per class $c\in\mathcal{C}$, ProxyNCA is then defined as 
\begin{equation}
    \mathcal{L}_{proxy} = -\frac{1}{b}\sum_{i\in\mathcal{B}}\log\left(  \frac{\exp(-d_e(\phi_i, \psi_{y_i})}{\sum_{c\in\mathcal{C}\setminus\{y_i\}} \exp(-d(\phi_i, \psi_{c})}  \right)
\end{equation}
%%%
\paragraph{Quadruplet~\cite{quadtruplet}}
The quadruplet loss is an extension to the triplet loss, which introduces higher level ordering constraint on sample embeddings. By using an anchor, a positive and two exclusive negatives, the quadruplet criterion is defined as:
\begin{equation}
    \mathcal{L}_{Quadr} = \sum_{\substack{i,j,k\in\mathcal{B}\\y_i=y_j, y_j\neq y_k}} \left[ d(\phi_i,\phi_j) - d(\phi_i,\phi_k) + \gamma_1 \right]_+ + \sum_{\substack{i,j,k,l\in\mathcal{B}\\y_i=y_j, y_j\neq y_k, y_l\neq y_k, y_l \neq y_j}} \left[ d(\phi_i,\phi_k) - d(\phi_l,\phi_k) + \gamma_2\right]_+
\end{equation}
with margin parameters $\gamma_1=1$ and $\gamma_2=0.5$. We utilize distance-weighted sampling to propose the first negative sample $k$, which we found to work better than the quadruplet sampling scheme originally proposed in the paper.
%%%
\paragraph{SNR~\cite{signal2noise}} 
The Signal-to-Noise-Ratio loss (SNR) introduces a novel distance metric based on the ratio between anchor embedding variance and variance of noise, which is simply defined as the difference between anchor and compared embedding. This optimises the embedding space directly for informativeness. The complete loss can then be written as
\begin{equation}
    \mathcal{L}_{SNR} = \sum_{i,j,k\in\mathcal{T}} \left[\frac{\sum_{m=1}^D (\phi_{i,m}-\phi_{j,m})}{\sum_{m=1}^D\phi_{i,m}^2} -
    \frac{\sum_{m=1}^D (\phi_{i,m}-\phi_{k,m})}{\sum_{m=1}^D\phi_{i,m}^2} + \gamma\right]_+ + \frac{\lambda}{b}\sum_{i\in\mathcal{B}}\left\Vert \sum_{m=1}^D\phi_{i,m}\right\Vert
\end{equation}
with margin parameter $\gamma=0.2$ and regularization $\lambda=0.005$ to ensure zero-mean distributions. Note that $\phi_{i,m}=\phi(x_i)_m$. 
%%%
\paragraph{SoftTriple~\cite{softriple}}
Similar to ProxyNCA, the SoftTriple objective function utilizes learnable data proxies to tackle the sampling problem. However, instead of class-discriminative proxies, a set of normalized intra-class proxies $\psi \in \Psi^c$ per class $c$ are learned using the NCA-based similarity measure $\mathcal{S}_i^c$ of a sample $i$ to all proxies of a class $c$. Denoting the set of available classes as $\mathcal{C}$ and the total set of proxies as $\Psi$, we get
\begin{align}
    \mathcal{S}_i^c      &= \sum_{\psi\in\Psi^c} \frac{\exp(\frac{1}{\gamma}\phi_i^T\psi)}{\sum_{\psi\in\Psi^c} \exp(\frac{1}{\gamma}\phi_i^T\psi)}\\
    \mathcal{L}_{STBase} &= -\frac{1}{b}\sum_{i\in\mathcal{B}}\log\frac{\exp(\lambda(\mathcal{S}_i^{y_i} - \delta))}{\exp(\lambda(\mathcal{S}_i^{y_i}-\delta)) + \sum_{y\in\mathcal{Y}\setminus \{y_i\}}\exp(\lambda\mathcal{S}_i^{y})}\\
    \mathcal{L}_{SoftTriple} &= \mathcal{L}_{STBase} + \tau\cdot\frac{\sum_{c\in\mathcal{C}}\sum_{\psi_1,\psi_2\in\Psi^c, \psi_1\neq \psi_2} \sqrt{2 - 2\psi_1^T\psi_2} }{|\mathcal{C}|\cdot|\Psi|\cdot(|\Psi|-1)}
\end{align}
The second term denotes a regularization on the learned proxies to ensure sparseness in the class set of proxies. For our tests, we utilised the following hyperparameter values (borrowing from the official implementation in \cite{softriple}): $\tau=0.2$, $\lambda=8$, $\delta=0.01$, $\gamma=0.1$ and the number of proxies per class $|\Psi^c|=2$ (higher values resulted in much worse performance). The proxy learning rate is set to $0.00001$.
%%%
\paragraph{Normalized Softmax~\cite{zhai2018classification}}
Similar to other classification-based losses in DML that are based on re-formulations of the standard softmax function (such as $\mathcal{L}_{arc}$), the normalized softmax loss is optimized by comparing input embeddings $\phi_i$ to class proxies $\psi\in\mathbb{R}^D$ per class $c\in\mathcal{C}$:
\begin{equation}
    \mathcal{L}_{NormSoft} = -\sum_{i\in\mathcal{B}}\log\left( \frac{\exp(\frac{\phi_i^T\psi_{y_i}}{T})}{\sum_{c\in\mathcal{C}\setminus\{y_i\}} \exp(\frac{\phi_i^T\psi_{c}}{T})}\right)
\end{equation}
with temperature $T=0.05$ for gradient boosting and class proxy learning rate set to $10^{-5}$.\\ 
%%%
\subsection{Tuple Mining}\label{supp:miner}
Basic contrastive, triplet or higher order ranking losses commonly need to mine their training tuples from the available mini-batch. In our study, we measure the influence of tuple sampling on the standard triplet loss, while utilising Distance-Weighted Mining for all ranking-based objective functions except N-Pair based methods.
\paragraph{Random Tuple Mining~\cite{face_verfication_inthewild}}
The trivial way involves the random sampling of tuples. Simply put, per sample $\{x_i\}_{i\in\mathcal{B}}$ we select a respective positive $\{j | y_j= y_i, i\neq j, j\in\mathcal{B}\}$ or negative sample $\{k |  y_j\neq y_i, i\neq k, k\in\mathcal{B}\}$. 
%%%
\paragraph{Semihard Triplet Mining~\cite{semihard}}
The potential number of triplets scales cubic in training set size. During learning, more and more of those triplets are correctly ordered and effectively provide no training signal \cite{semihard}, thus impairing the remaining training process. To alleviate this, negative samples are carefully selected based on the anchor-positive sample distance (which are sampled at random). Given an anchor embedding $\phi_a$ and its positive $\phi_p$, the negative is sampled randomly from the set 
\begin{equation}
    \phi_n \in \{\phi_n | n\in\mathcal{B}, y_n\neq y_a, \left\Vert\phi_a - \phi_p\right\Vert_2^2 < \left\Vert\phi_a - \phi_n\right\Vert_2^2\}. 
\end{equation}
This way, only negatives are considered which are reasonably hard to separate from an anchor. Moreover, this mining strategy avoids the sampling of overlay hard negatives, which often correspond to data noise and potentially lead to model collapses and bad local minima~\cite{semihard}.
%%%
\paragraph{Softhard Triplet Mining~\cite{rothgithub}}
While it was justifiably noted in \cite{semihard} that a selection of 'hard' samples hurts training, \cite{rothgithub} show that a probabilistic (soft) selection of potentially hard candidates can actually benefit performance. Given an anchor embedding $\phi_a$, positive $\phi_p$ and $\phi_n$ are randomly selected from 
\begin{equation}
    \phi_n \in \{\phi_n | n\in\mathcal{B}, y_n\neq y_a, \left\Vert\phi_a - \phi_n\right\Vert_2^2 < \argmax_{p\in\mathcal{B}, y_a=y_p} \left\Vert\phi_a - \phi_p\right\Vert_2^2\}
\end{equation}
and
\begin{equation}
    \phi_a \in \{\phi_a | a\in\mathcal{B}, y_n= y_a, \left\Vert\phi_a - \phi_n\right\Vert_2^2 > \argmin_{n\in\mathcal{B}, y_a\neq y_p} \left\Vert\phi_a - \phi_n\right\Vert_2^2\}.
\end{equation}
Doing so provides a selection of 'hard' positivies and negatives. This reduces the risk of potential model collapses and bad local minima (as noted in \citet{semihard}).
%%%
\paragraph{Distance-Weighted Tuple Mining~\cite{margin}} 
In DML, the embedding spaces are typically normalized to a $D$-dimensional (unit) hypersphere $\mathbb{S}^{D-1}$ for regularisation purposes \cite{margin}. The analytical distribution of pairwise distances on a hypersphere follows
\begin{equation}
    q(d_e(\phi_i,\phi_j)) \propto d_e(\phi_i,\phi_j)^{D-2}[1-\frac{1}{4}d_e(\phi_i,\phi_j)]^{\frac{D-3}{2}}
\end{equation}

for arbitrary embedding pairs $\phi_i, \phi_j\in\mathbb{S}^{D-1}$. In order to sample negatives from the whole range of possible distances to an anchor, \citet{margin} propose to sample negatives based on a distance distribution inverse to $q$, i.e.
\begin{equation}
    P(n|a) \propto \min(\lambda, q^{-1}(d_e(\phi_a, \phi_n))
\end{equation}
We set $\lambda=0.5$ and limit the distances to $1.4$.
%%%%%

\subsection{Evaluation Metrics}\label{supp:metrics}
In this section, we examine the evaluation metrics to measure the performance of the studied models on a the testset $\mathcal{X_{\text{test}}}$. 
\paragraph{Recall@k~\cite{recall}} Let 
\begin{equation}\label{eq:fk}
\mathcal{F}_q^k = \argmin_{\mathcal{F}\subset\mathcal{X_{\text{test}}},|\mathcal{F}|=k} \sum_{x_f\in\mathcal{F}} d_e(\phi(x_q),\phi(x_f)) 
\end{equation}
be the set of the first $k$ nearest neighbours of a sample $x_p$, then we measure Recall@k as 

\begin{equation}
    R@k = \frac{1}{|\mathcal{X_{\text{test}}}|}\sum_{x_q\in\mathcal{X_{\text{test}}}}  
                \begin{cases}
                1 \qquad \exists \; x_i \in \mathcal{F}_q^k \; \text{s.t.} \; y_i = y_q\\
                0 \qquad \text{otherwise}
                \end{cases}
\end{equation}

which measures the average number of cases in which for a given query $x_q$ there is at least one sample among its top $k$ nearest neighbours $x_i$ with the same class, i.e. $y_i = y_q$.
\paragraph{Normalized Mutual Information (NMI)~\cite{nmi}} To measure the clustering quality using NMI, we embed all samples $x_i \in \mathcal{X_{\text{test}}}$ to obtain $\Phi_{\mathcal{X_{\text{test}}}}$ and perform a clustering (e.g. $K$-Means \cite{kmeans}). Following, we assign all samples $x_i$ a cluster label $w_i$ indicating the closest cluster center and define $\Omega = \{\omega_k\}_{k=1}^K$ with $\omega_k = \{i | w_i=k\}$ and $K = |\mathcal{C}|$ being the number of classes and clusters. Similarly for the true labels $y_i$ we define $\Upsilon = \{\upsilon_c\}_{c=1}^K$ with $\upsilon_c = \{i | y_i=c\}$. The normalized mutual information is then computed as 
\begin{equation}
    NMI(\Omega,\Upsilon) = \frac{I(\Omega,\Upsilon)}{2(H(\Omega) + H(\Upsilon)}
\end{equation}
with mutual Information $I(\cdot,\cdot)$ between cluster and labels, and entropy $H(\cdot,\cdot)$ on the clusters and labels respectively.
%%%
\paragraph{F1-Score~\cite{npairs}} The F1-score measures the harmonic mean between precision and recall and is a commonly used retrieval metric, placing equal importance to both precision and recall. It is defined as 
\begin{equation}
    F1 = \frac{2PR}{P+R}
\end{equation}
with precision $P$ and Recall $R$ defined over nearest neighbour retrieval as done for Recall@k.
%%%
\paragraph{Mean Average Precision measured on Recall (mAP) @C and @1000:} The mAP-score measured on recall follows the same definition as standard mAP, however the recalled samples are determined by the nearest neighbour ranking. In our case, we propose to use mAP@C and mAP@C. mAP@C is equivalent to the mean over the class-wise average precision@$k_c$ with $k_c$ being the number of samples with label $c \in \mathcal{C}$, which only recalls as many samples as there are members in a class. For completeness, we also use mAP@1000, which recalls the 1000 nearest samples. With $\mathcal{F}^{k_c}_q$ defined as in eq. \ref{eq:fk}, this gives

\begin{equation}
    \text{mAP@C} = \frac{1}{|\mathcal{X}_{\text{test}}|}\sum_{c\in\mathcal{C}} \sum_{x_q\in \mathcal{X}_{\text{test}} \wedge y_q = c} \frac{\left|\{x_i \in \mathcal{F}^{k_c}_q | y_i=y_q \}\right|}{k_c}
\end{equation}

\begin{equation}
    \text{mAP@1000} = \frac{1}{|\mathcal{X}_{\text{test}}|}\sum_{c\in\mathcal{C}} \sum_{x_q\in \mathcal{X}_{\text{test}} \wedge y_q = c} \frac{\left|\{x_i \in \mathcal{F}^{1000}_q | y_i=y_q \}\right|}{1000}
\end{equation}

%%%%%%%%%%%%%%%%%%%%%%%%%%%%%%%%%%%%%%%%%%%%%%%%%%%%%%%%%%%%%%%%%%%%
\section{Evaluating the robustness of $p_\text{switch}$ choices}\label{supp:switch_ablate}
For all benchmarks, we evaluate the change in performance for varying values of $p_\text{switch}$. As can be seen in figure \ref{fig:rand_push_hyperpar}, performance steadily increases up to a point of saturation, where the regulatory signal overshadows the discriminative one. In addition, its clear that performance increases are reasonably robust against changes in $p_\text{switch}$.
\input{figures/random_pushop_abl}

%%%%%%%%%%%%%%%%%%%%%%%%%%%%%%%%%%%%%%%%%%%%%%%%%%%%%%%%%%%%%%%%%%%
\section{Visual comparison of DML objectives on benchmarks}\label{supp:visualsummary}
\input{figures/full_first_page}

This figure is the full version of the first page figure (Fig. \ref{fig:fp1}). It qualitatively supports the saturation of performance noted in section \ref{sec:eval_ana} and table \ref{tab:final_results}.

%%%%%%%%%%%%%%%%%%%%%%%%%%%%%%%%%%%%%%%%%%%%%%%%%%%%%%%%%%%%%%%%%%%%
\section{Correlation between performance and spectral decay $\rho$}
\input{figures/dists_to_perf}

Similar to Fig. 5 (rightmost) in the main paper, we now provide a more detailed illustration in Fig. \ref{fig:regularization_effect} comparing the performance of the training objectives and their corresponding spectral decay $\rho(\Phi)$. For ranking losses, we further include the results using $\rho$-regularization while training, which further shows that in each case a gain in performance is related to a decrease of $\rho(\Phi)$. Especially the contrastive loss~\cite{contrastive} greatly profits from our proposed regularization, as also indicated by the analysis of the singular value spectra (cf. Fig. 8 of main paper). Its large gains, more then $5\%$ on the CARS196 dataset, is well explained by comparison of its training objective with those of triplet-based formulations. The latter optimizes over relative positive ($d_\phi(x_a,x_p)$)) and negative distances ($d_\phi(x_a,x_n)$) up to a fixed margin $\gamma$, which counteracts a compression of the embedding space to a certain extend. On the other hand, the constrastive loss, while controlling only the negative distances by $\gamma$, is able to perform an unconstrained contraction of entire classes, which facilitates overly compressed embedding spaces $\Phi$.

%%%%%%%%%%%%%%%%%%%%%%%%%%%%%%%%%%%%%%%%%%%%%%%%%%%%%%%%%%%%%%%%%%%%
\section{Analysis of per-class singular value spectra}\label{supp:svd_progression}
\input{figures/per_class_singular_vector_progression}
In Sec. 5 of our main paper we analyze generalization in DML by considering the decay of the singular value spectrum over all embedded samples $\Phi_\mathcal{X}$. Thus, we analyze the general compression of the entire embedding space $\Phi$ as unseen test classes can be projected anywhere in $\Phi$, in contrast to \citet{manifoldmixup} which conduct a class-conditioned analysis for i.i.d. classification problems. In order to show that the effect of $\rho$-regularization (as shown in Fig. 8 in main paper) is also reflected in the class-conditioned singular value spectrum, we perform SVD on $\Phi_{y_l}$ and subsequently average over all classes $y_l \in \mathcal{Y}$. Fig. \ref{fig:per_class_singular_value_progression} compares the sorted, first $35$ singular values for both, models trained with and without $\rho$-regularization. We clearly see that the regularization decreases the average decay of singular values similar to the total singular value spectra shown in the main paper.

%%%%%%%%%%%%%%%%%%%%%%%%%%%%%%%%%%%%%%%%%%%%%%%%%%%%%%%%%%%%%%%%%%%%
\section{Comparison to state-of-the-art approaches on SOP dataset}\label{supp:sop_sota}
\input{tables/sota_sop}

In this section we provide a detailed comparison between current state-of-the-art DML approaches and our strongest baseline model, margin loss (D, $\beta=1.2$)~\cite{margin}, on the SOP dataset in Tab. \ref{tab:eval_sop}. The results for these approaches are taken from their public manuscripts. We observe that our baseline model outperforms each of the models using varying architectures, but especially other ResNet50-based implementations. While R50 proves to be a stronger base network (cf. Fig. 2 of main paper) than GoogLeNet based model, improvements over MIC and D\&C using the same backbone by at least $0.9\%$ and methods based on the similarly strong Inception-BN showcase the relevance of a well-defined baseline. Additionally, even though Rank and ABE employ considerable more powerful network ensembles, our carefully motivated baseline exhibits competitive performance.

%%%%%%%%%%%%%%%%%%%%%%%%%%%%%%%%%%%%%%%%%%%%%%%%%%%%%%%%%%%%%%%%%%%%
\section{2D Toy Examples}\label{supp:more_toy}
For our toy examples, we use a fully-connected network with two 30 neuron layers. Both input and embedding dimension are 2D, while the latter is normalized onto the unit circle. Each of the four training and test lines contain 15 samples taken from either the diagonal or vertical/horizontal line segments, respectively. We train the networks both with and without regularization for $200$ iterations, a batchsize of $24$ and learning rate of $0.03$ using a standard contrastive loss (eq. \ref{eq:contrastive}) with margin $\gamma=0.1$. For regularisation, we set $p_{\text{switch}}=0.001$.
Similar to Fig.6 in the main paper, Fig. \ref{fig:straight_toy_example} shows another 2D toy example based on vertical lines which again demonstrates the effect of compression and of our proposed $\rho$-regularization. The example consists of four training lines that are separable only by their $x$-coordinate and a test set of lines which are separable by their $y$-coordinate. As we observe, the test samples are collapsed onto a single point in the non-regularized embedding space, thus can not be distinguished. In contrast, the regularized representation allows us to separate the test classes and, further, exhibits a decreased decay in the singular value spectrum.
\input{figures/straight_toy_ex}

%%%%%%%%%%%%%%%%%%%%%%%%%%%%%%%%%%%%%%%%%%%%%%%%%%%%%%%%%%%%%%%%%%%%
\section{Influence of Manifold Mixup on DML}\label{supp:mmanifold}
Now, we examine the effect of applying the regularization proposed in ManifoldMixup~\cite{manifoldmixup} on the DML transfer learning setting. As ManifoldMixup has been proposed to increase the compression of a learned representation in the context of standard supervised classification, it is expected to decrease the performance of DML models. For that, we train three different DML models on the CUB200-2011 dataset: (1) Normalized Softmax, (2) Triplet with Distance Sampling and (3) Margin loss with $\beta=0.6$ and Distance Sampling. For (1), the implementation directly follows the standard implementation noted in \citet{manifoldmixup}. For the ranking-based training objectives, we perform mixup in our ResNet50 and generate the mixed class labels, which consequently have either one (if image from the same class are mixed) or two entries (if images from different classes are mixed). Per (mixed) anchor embedding, this gives rise to up to two possible sets of triplets, for which we compute the loss and weigh it by the respective mixup coefficient $\lambda_k$:
\begin{equation}
    \mathcal{L}_{tripl}^{\text{Mix}} = \frac{1}{b} \sum_{k=1}^2 \sum^{b}_{\substack{(a,p,n)\in\mathcal{T}^k \\ y_a=y_p\neq y_n}} \lambda_k\cdot\left[d_e(\phi^\lambda_a, \phi^\lambda_p) - d_e(\phi^\lambda_a, \phi^\lambda_n) + \gamma\right]_+
\end{equation}
where $\mathcal{T}^k$ denotes the set of triplets given the $k$-th mixup class-label entry and $\lambda_k$ the respective interpolation value. We use the notation $\phi^\lambda_x$ to denote that we now operate on mixup embeddings. For training, we use the standard hyperparameters as described in Sec. 4.1. of the main paper and a mixup-$\alpha$ of $2$ to sample the interpolation values $\lambda\sim\beta(\alpha,\alpha)$ (see \citet{manifoldmixup}).\\
The results after rerunning baselines and mixup-variants are shown in Fig. \ref{fig:mixup_study}. As expected, applying ManifoldMixup leads to more compressed representations (indicated by a stronger spectral decay and lower $\rho(\Phi)$-scores) at the cost of reduced generalization performance. This holds both for the spectrum across the fully embedded dataset as well as on a class level.

\input{figures/mixup_study}

%%%%%%%%%%%%%%%%%%%%%%%%%%%%%%%%%%%%%%%%%%%%%%%%%%%%%%%%%%%%%%%%%%%%
\section{Detailed Results}\label{supp:detailed}
This section contains detailed results per method and evaluation metric for the method comparisons (Tab. 2 in the main paper) and evaluation of batch-creation methods (Fig. 3 in main paper). The resp. tables for the method comparisons are Tab. \ref{tab:eval_cub} for CUB200-2011 \cite{cub200-2011}, Tab. \ref{tab:eval_car} for CARS196 \cite{cars196} and Tab. \ref{tab:eval_sop} for Stanford Online Products (SOP) \cite{lifted}. In addition, the switch probability $p_{\text{switch}}$ for each regularised method is noted as well. The batch-creation methods are evaluated in detail in Tab. \ref{tab:meta_sampling_cub} for CUB200-2011, Tab. \ref{tab:meta_sampling_cars} for CARS196 and Tab. \ref{tab:meta_sampling_sop} for SOP. 
\input{tables/eval_baselines.tex}

\input{tables/total_comparison.tex}

%% file: figures/random_pushop_abl.tex
\begin{figure}[t]
\begin{center}
\includegraphics[width=0.8\linewidth]{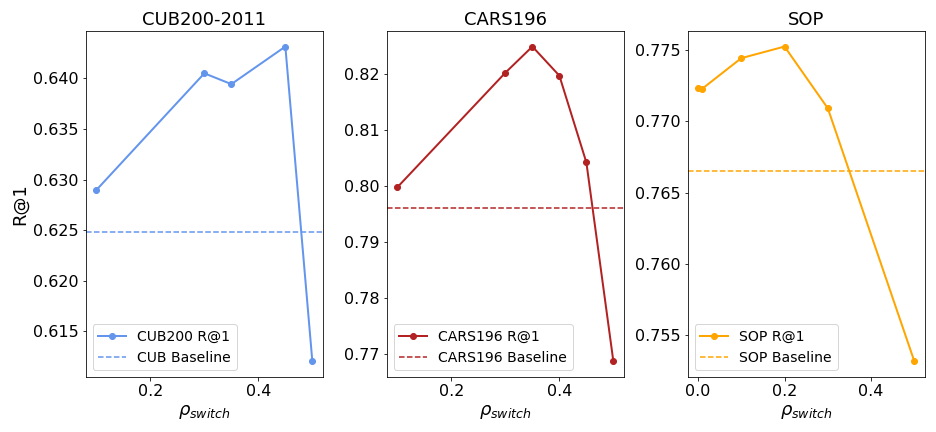}
\end{center}
\vspace{-8pt}
   \caption{\textit{Analysis of the influence of} $p_{\text{switch}}$ on Recall@1 using margin loss with $\beta=0.6$. Dashed lines denote performance without $\rho$-regularization.}
\label{fig:rand_push_hyperpar}
\vspace{-8pt}
\end{figure}

% Correlation between performance and push-operation/Total Cluster Density Ratio (TD-Ratio) for CUB200-2011 and CARS196. Both have a similar dataset structure. Generally, we found that optimizing within the interval [0,0.45] to be best.

%% file: figures/full_first_page.tex
\begin{figure}[t]
\centering
%\begin{subfigure}[c]{1\linewidth}
\begin{center}
\includegraphics[width=0.99\linewidth]{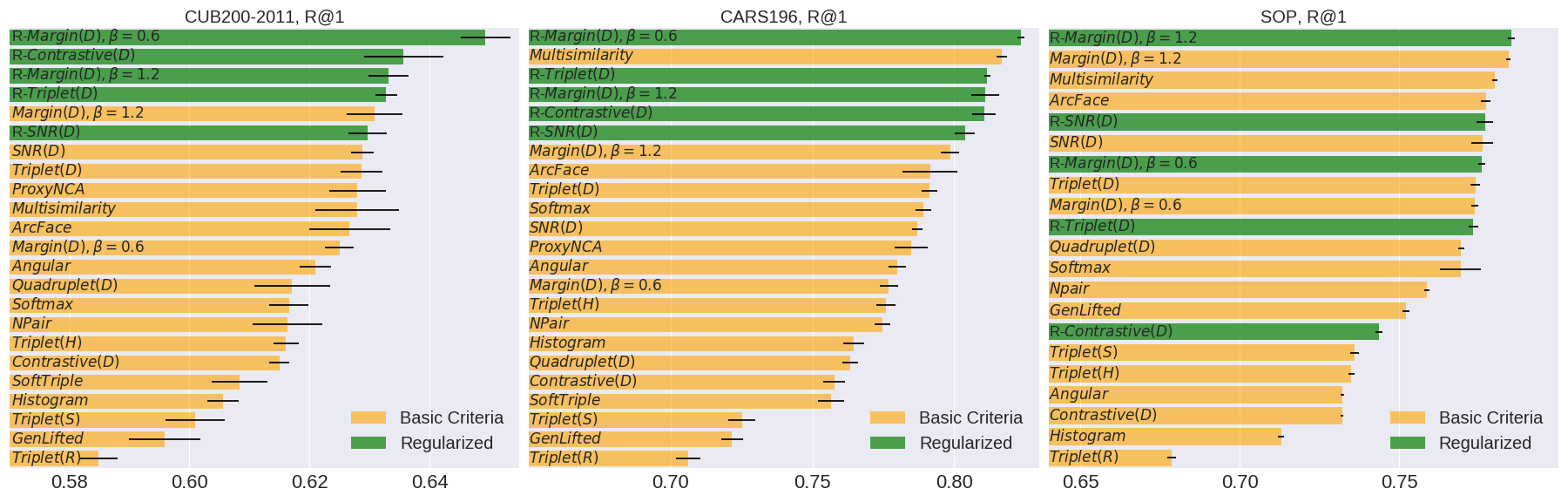}
\end{center}
\vspace{-5pt}
   \caption{\textit{Mean recall performance and standard deviation} of various DML objective functions trained with (green) and without (orange) our proposed regularization.}
\label{fig:fp_full}
%\end{subfigure}
\vspace{-12pt}
\end{figure}

%% file: figures/dists_to_perf.tex
\begin{figure*}[h]
\begin{subfigure}[c]{1\linewidth}
\begin{center}
\includegraphics[width=1\linewidth]{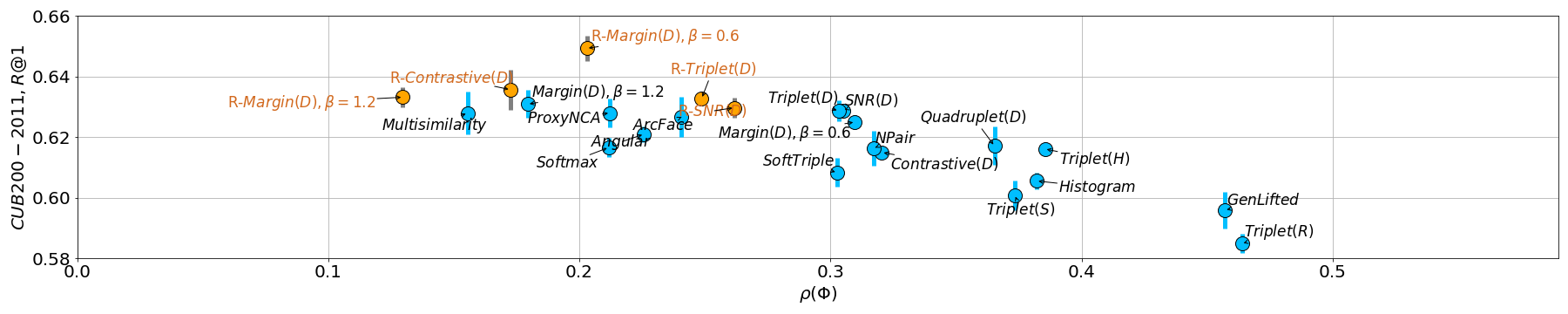}
\end{center}
\end{subfigure}
\begin{subfigure}[c]{1\linewidth}
\begin{center}
\includegraphics[width=1\linewidth]{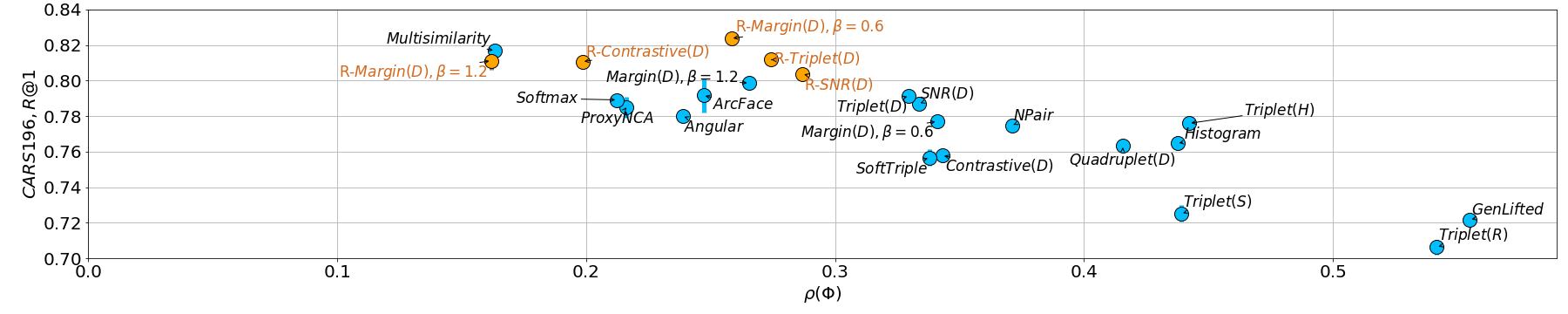}
\end{center}
\end{subfigure}
\begin{subfigure}[c]{1\linewidth}
\begin{center}
\includegraphics[width=1\linewidth]{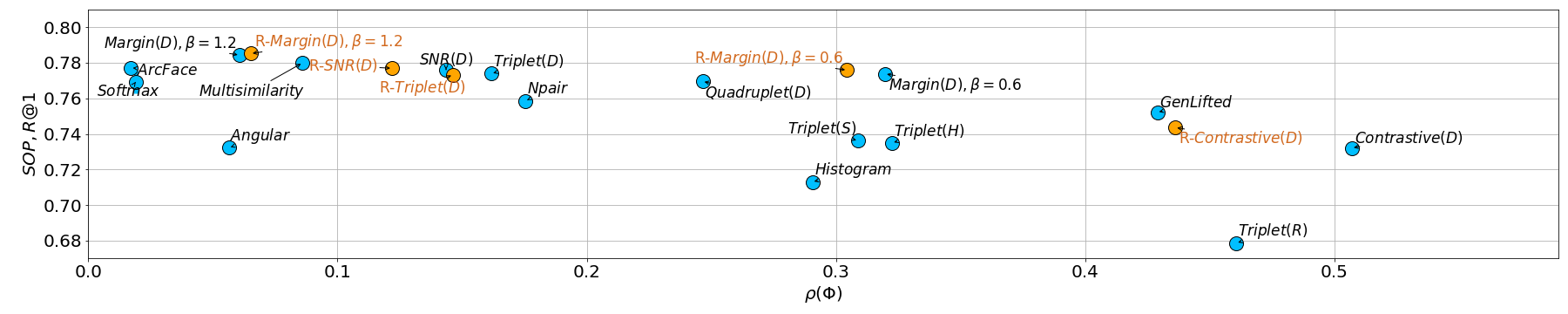}
\end{center}
\end{subfigure}
\vspace{-8pt}
   \caption{Relation between $\rho(\Phi)$ and generalization performance on Recall@1 for models trained with (orange) and without (blue) $\rho$-regularization. We report mean results and error-bars (gray). When error is small, bars are covered.}
\label{fig:regularization_effect}
\vspace{-8pt}
\end{figure*}

% Linear fit (blue) was computed with basic criteria (gray) and errors, excluding the anticollapse criteria (green). Exact Values can be taken table .ref. Especially looking at triplet loss with varying sampling methods, one can see the direct influence of reduces packing density on a specific loss function achieved through different sampling methods.

%% file: figures/per_class_singular_vector_progression.tex
\begin{figure}[h]
\begin{center}
\includegraphics[width=0.8\linewidth]{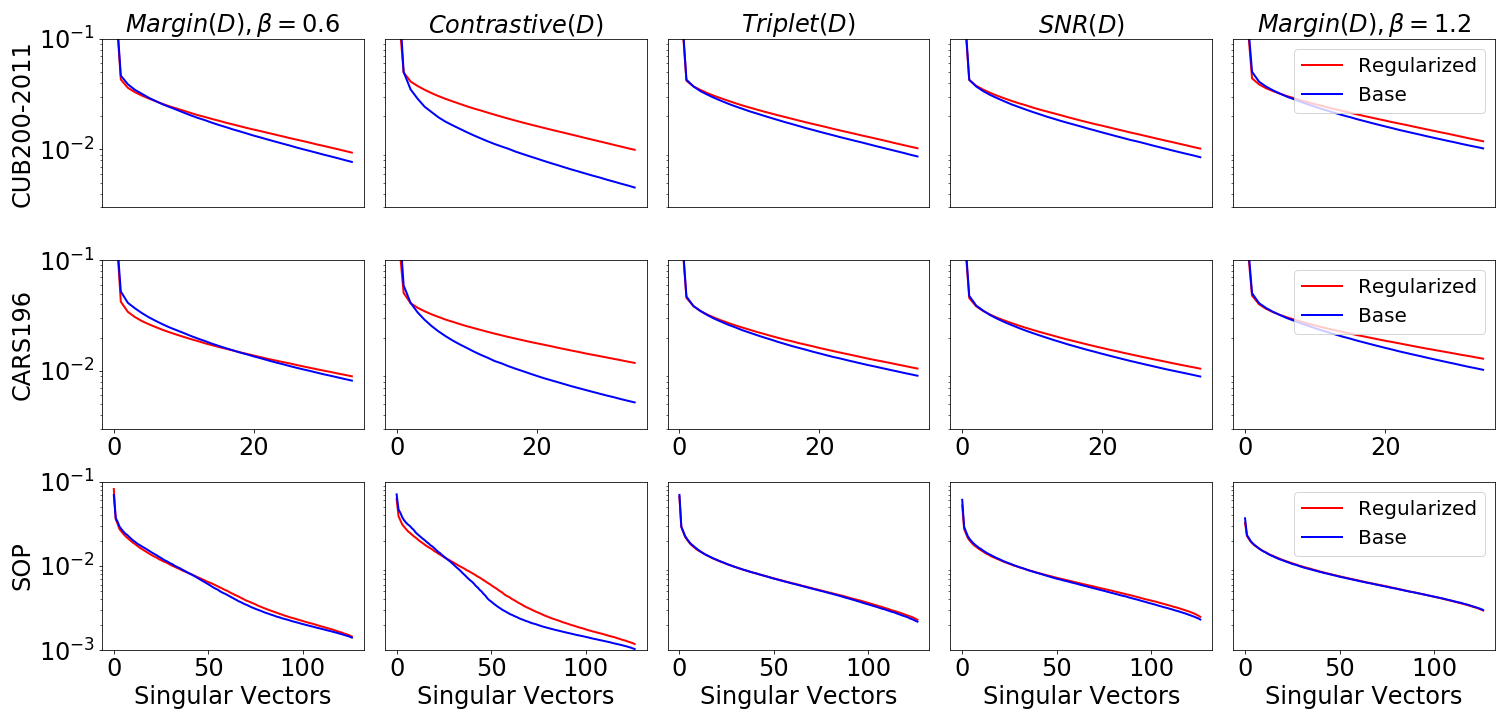}
\end{center}
   \caption{Averaged class-conditioned spectra of singular values for models trained with (red) and without (blue) $\rho$-regularization for various ranking-based loss functions.}
\label{fig:per_class_singular_value_progression}
\end{figure}

%% file: tables/sota_sop.tex
\begin{table*}[t]
    \setlength\tabcolsep{1.5pt}
    \centering
     \begin{tabular}{l|c||c|ccc|c}
        \toprule
        Approach & Architecture & Dim & R@1 & R@10 & R@100  & NMI \\
        \midrule
        DVML\cite{dvml} & GoogLeNet    & 512 & 70.2 & 85.2 & 93.8 & 90.8\\
        HTL\cite{htl}   & Inception-BN & 512 & 74.8 & 88.3 & 94.8  & -\\
        MIC\cite{mic}   & ResNet50     & 128 & 77.2 & 89.4 & 95.6 & 90.0\\
        D\&C\cite{Sanakoyeu_2019_CVPR} & ResNet50& 128 & 75.9 & 88.4 & 94.9 & 90.2\\
        \midrule
        Rank\cite{rankedlist} & Inception-BN & 1536 & \textbf{79.8} & \textbf{91.3} & \textbf{96.3} & 90.4\\
        ABE\cite{abe}         & GoogLeNet    & 512 & 76.3 & 88.4 & 94.8 & -\\
        \midrule
        \midrule
        \textbf{Margin (ours)}\cite{margin} & ResNet50 & 128 & 78.4 & - & - & 90.4\\
        \bottomrule
    \end{tabular}
    \caption{Comparison to the state-of-the-art DML methods on SOP\cite{lifted}. \emph{Dim} denotes the dimensionality of $\phi_\theta$.}
    %As can be seen, our adaptive sampling offers state-of-the-art results on CUB200-2011 and competitive results on CARS196 \& SOP while using a standard triplet-based marginloss setup. For completeness, we also list results achieved by ensemble methods or those with significant architecture changes.} % * indicates our ResNet50 implementation.}
    \label{tab:sota}
\end{table*}

%% file: figures/straight_toy_ex.tex
\begin{figure*}[h]
\begin{center}
\includegraphics[width=1\linewidth]{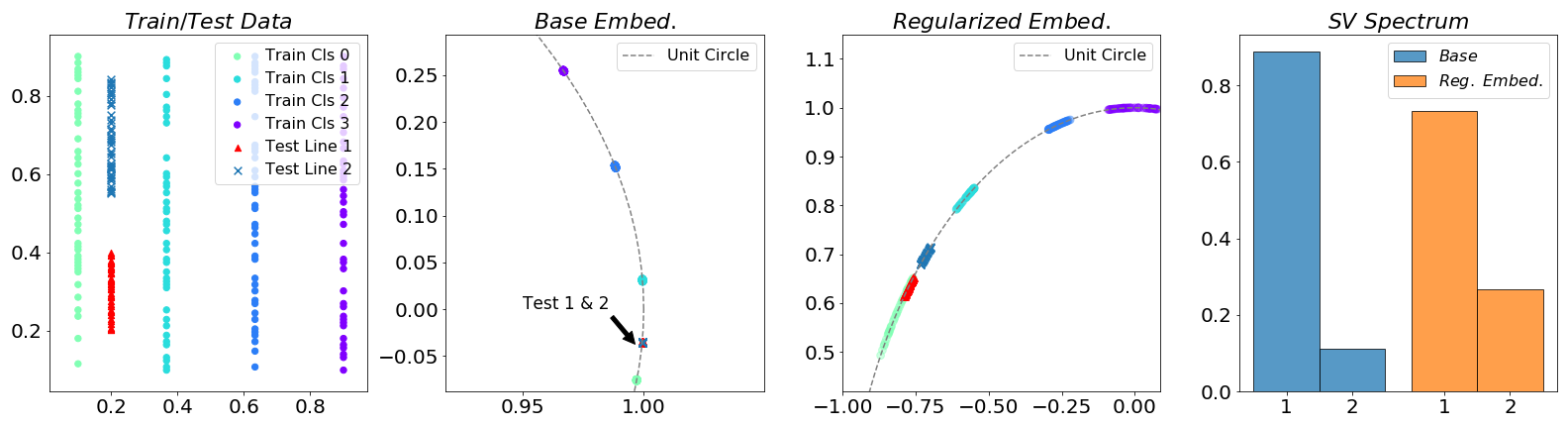}
\end{center}
   \caption{Toy example based on horizontally discriminative training data, where to goal is to generalize to vertically discriminative test data. (Leftmost) training and test data. (Mid-left) A small, normalized two-layer fully-connected network trained with standard contrastive loss fails to separate both test classes as it never has to utilize vertical discrimination. (Mid-right) The regularized embedding successfully separates the test classes by introducing additional features and decreasing the spectral decay. (Rightmost) Singular value spectra of training embeddings learned with and without regularization.}
\label{fig:straight_toy_example}
\end{figure*}

%% file: figures/mixup_study.tex
\begin{figure}[h]
\centering
\begin{subfigure}[c]{0.8\linewidth}
\begin{center}
\includegraphics[width=1\linewidth]{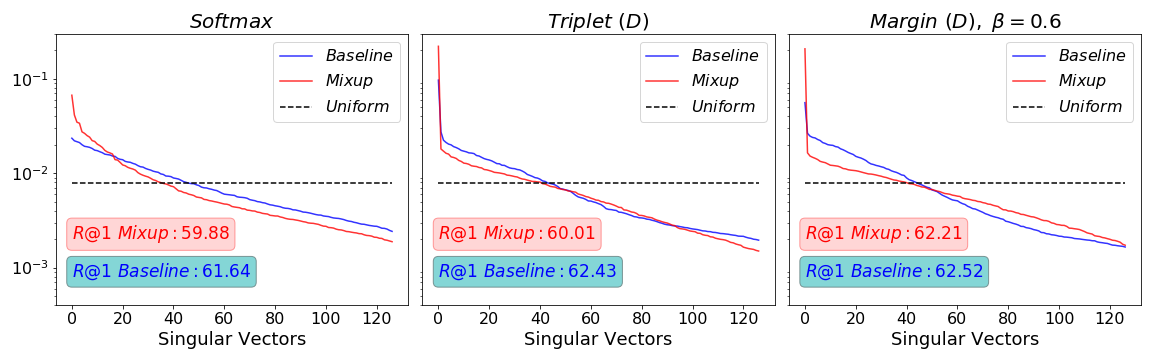}
\caption{Singular Value Spectrum for all embeddings}
\end{center}
\end{subfigure}
\begin{subfigure}[c]{0.8\linewidth}
\begin{center}
\includegraphics[width=1\linewidth]{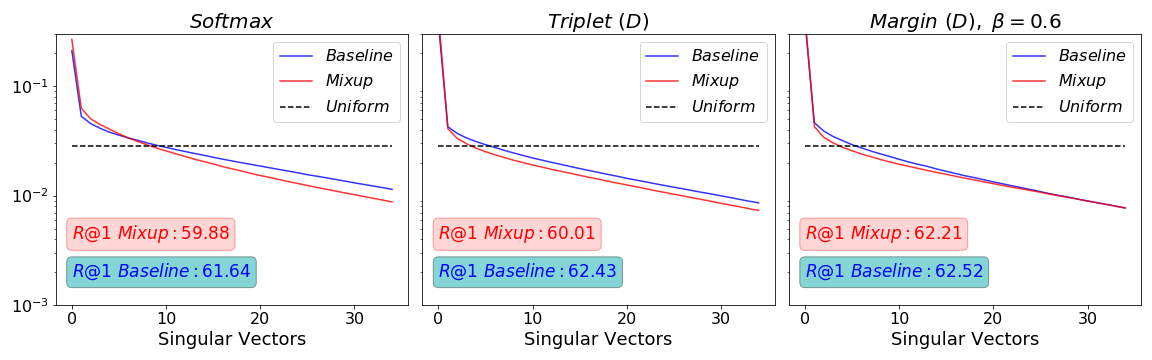}
\caption{Singular Value Spectrum for per-class embeddings}
\end{center}
\end{subfigure}
\caption{Evaluation of Mixup Influence on zero-shot generalization under heavy distribution shift.}
\label{fig:mixup_study}
\end{figure}

%% file: tables/eval_baselines.tex
\begin{table*}[h]
 \small
   \setlength\tabcolsep{1.4pt}
   \centering
   %\begin{tabular}{l|c|ccc|c}
   \begin{tabular}{l|c|c|c|c|c|c}
     \toprule
     \multicolumn{7}{c}{CUB200-2011\cite{cub200-2011}} \\
     \midrule
     Approach & R@1 & R@2 & F1 & mAP@C & mAP@1000 & NMI\\
     \midrule
    Imagenet & $43.77$ & $57.56$ & $19.14$ & $9.48$ & $14.62$ & $52.91$\\
    \midrule     
    Angular & $62.10\pm0.27$ & $73.68\pm0.39$ & $37.53\pm0.13$ & $22.06\pm0.20$ & $30.56\pm0.27$ & $67.59\pm0.26$\\
    ArcFace & $62.67\pm0.67$ & $74.38\pm0.25$ & $37.33\pm0.51$ & $23.05\pm0.43$ & $31.62\pm0.68$ & $67.66\pm0.38$\\
    Contrastive (D) & $61.50\pm0.17$ & $72.95\pm0.32$ & $35.40\pm0.75$ & $23.46\pm0.18$ & $32.38\pm0.21$ & $66.45\pm0.27$\\
    GenLifted & $59.59\pm0.60$ & $71.63\pm0.38$ & $34.86\pm0.16$ & $22.03\pm0.14$ & $30.30\pm0.13$ & $65.63\pm0.14$\\
    Histogram & $60.55\pm0.26$ & $72.08\pm0.20$ & $33.88\pm0.56$ & $22.65\pm0.20$ & $31.24\pm0.27$ & $65.26\pm0.23$\\
    Multisimilarity & $62.80\pm0.70$ & $74.37\pm0.52$ & $39.03\pm0.63$ & $22.58\pm0.37$ & $30.92\pm0.49$ & \blue{$\mathbf{68.55\pm0.38}$}\\
    Margin (D, $\beta=0.6$) & $62.50\pm0.24$ & $74.15\pm0.33$ & $36.34\pm0.61$ & $23.83\pm0.20$ & $32.90\pm0.29$ & $67.02\pm0.37$\\
    Margin (D, $\beta=1.2$) & $63.09\pm0.46$ & $74.41\pm0.37$ & $38.36\pm0.66$ & $23.61\pm0.31$ & $32.63\pm0.40$ & $68.21\pm0.33$\\
    NPair & $61.63\pm0.58$ & $73.33\pm0.42$ & $37.29\pm0.42$ & $22.16\pm0.29$ & $30.83\pm0.30$ & $67.64\pm0.37$\\
    ProxyNCA & $62.80\pm0.48$ & $74.03\pm0.15$ & $36.20\pm0.73$ & $23.94\pm0.37$ & $33.13\pm0.59$ & $66.93\pm0.38$\\
    Quadruplet (D) & $61.71\pm0.63$ & $73.26\pm0.33$ & $35.74\pm0.62$ & $23.20\pm0.24$ & $31.87\pm0.29$ & $66.60\pm0.41$\\
    SNR (D) & $62.88\pm0.18$ & $74.33\pm0.26$ & $36.91\pm0.42$ & $23.48\pm0.14$ & $32.24\pm0.21$ & $67.16\pm0.25$\\
    SoftTriple & $60.83\pm0.47$ & $71.61\pm0.58$ & $32.16\pm0.50$ & $22.43\pm0.29$ & $31.33\pm0.29$ & $64.27\pm0.36$\\
    Softmax & $61.66\pm0.33$ & $73.31\pm0.39$ & $35.94\pm0.59$ & $22.19\pm0.20$ & $30.67\pm0.26$ & $66.77\pm0.36$\\
    \midrule
    Triplet (D) & $62.87\pm0.35$ & $74.31\pm0.28$ & $37.30\pm0.32$ & $23.59\pm0.12$ & $32.64\pm0.13$ & $67.53\pm0.14$\\
    Triplet (R) & $58.48\pm0.31$ & $70.51\pm0.24$ & $31.95\pm0.44$ & $21.12\pm0.10$ & $29.08\pm0.12$ & $63.84\pm0.30$\\
    Triplet (S) & $60.09\pm0.49$ & $71.75\pm0.27$ & $34.46\pm0.54$ & $22.49\pm0.26$ & $31.40\pm0.39$ & $65.59\pm0.29$\\
    Triplet (H) & $61.61\pm0.21$ & $72.94\pm0.34$ & $35.10\pm0.37$ & $22.63\pm0.23$ & $31.22\pm0.27$ & $65.98\pm0.41$\\
    \midrule
    \midrule
    R-Contrastive (D) & $63.57\pm0.66$ & $74.57\pm0.63$ & $37.70\pm0.53$ & $23.54\pm0.37$ & $32.40\pm0.66$ & $67.63\pm0.31$\\
    R-Margin (D, $\beta=0.6$) & \blue{$\mathbf{64.93\pm0.42}$} & \blue{$\mathbf{75.58\pm0.25}$} & \blue{$\mathbf{38.93\pm0.54}$} & \blue{$\mathbf{24.11\pm0.20}$} & \blue{$\mathbf{33.38\pm0.27}$} & $68.36\pm0.32$\\
    R-Margin (D, $\beta=1.2$) & $63.32\pm0.33$ & $74.80\pm0.27$ & $38.09\pm0.97$ & $22.80\pm0.46$ & $31.44\pm0.67$ & $67.91\pm0.66$\\
    R-SNR (D) & $62.97\pm0.32$ & $74.66\pm0.06$ & $38.25\pm0.41$ & $23.13\pm0.23$ & $32.23\pm0.47$ & $68.04\pm0.34$\\
    R-Triplet (D) & $63.28\pm0.18$ & $74.97\pm0.28$ & $38.03\pm0.77$ & $23.28\pm0.28$ & $32.36\pm0.47$ & $67.86\pm0.51$\\

    \bottomrule
    \end{tabular}
    \caption{Comparison of DML setups for CUB200-2011. We report all relevant performance metrics. Training is done over 150 epochs.}
    \label{tab:eval_cub}
 \end{table*}

 \begin{table*}[h]
 \small
   \setlength\tabcolsep{1.4pt}
   \centering
   %\begin{tabular}{l|c|ccc|c}
   \begin{tabular}{l|c|c|c|c|c|c}
     \toprule
     \multicolumn{7}{c}{CARS196\cite{cars196}} \\
     \midrule
     Approach & R@1 & R@2 & F1 & mAP@C & mAP@1000 & NMI\\
     \midrule
    Imagenet & $36.39$ & $48.11$ & $8.90$ & $4.03$ & $6.92$ & $37.96$\\
    \midrule
    Angular & $78.00\pm0.32$ & $85.97\pm0.18$ & $36.40\pm0.75$ & $22.18\pm0.35$ & $29.29\pm0.37$ & $66.48\pm0.44$\\
    ArcFace & $79.16\pm0.97$ & $87.02\pm0.54$ & $36.36\pm1.97$ & $23.44\pm0.68$ & $31.36\pm0.99$ & $66.99\pm1.08$\\
    Contrastive (D) & $75.78\pm0.39$ & $84.17\pm0.27$ & $33.10\pm0.63$ & $23.19\pm0.33$ & $30.43\pm0.51$ & $64.04\pm0.13$\\
    GenLifted & $72.17\pm0.38$ & $81.94\pm0.30$ & $32.46\pm0.43$ & $21.66\pm0.24$ & $28.66\pm0.31$ & $63.75\pm0.35$\\
    Histogram & $76.47\pm0.38$ & $84.50\pm0.36$ & $33.04\pm0.67$ & $23.21\pm0.10$ & $30.31\pm0.14$ & $64.15\pm0.36$\\
    Multisimilarity & $81.68\pm0.19$ & $88.86\pm0.14$ & $40.95\pm0.72$ & $24.22\pm0.27$ & $31.92\pm0.44$ & \blue{$\mathbf{69.43\pm0.38}$}\\
    Margin (D, $\beta=0.6$) & $77.70\pm0.32$ & $85.67\pm0.19$ & $35.04\pm0.49$ & $24.08\pm0.27$ & $31.99\pm0.29$ & $65.29\pm0.32$\\
    Margin (D, $\beta=1.2$) & $79.86\pm0.33$ & $87.46\pm0.20$ & $38.44\pm0.64$ & $24.72\pm0.21$ & $32.50\pm0.28$ & $67.36\pm0.34$\\
    NPair & $77.48\pm0.28$ & $85.73\pm0.18$ & $35.88\pm0.40$ & $23.15\pm0.25$ & $30.70\pm0.14$ & $66.55\pm0.19$\\
    ProxyNCA & $78.48\pm0.58$ & $86.20\pm0.50$ & $34.66\pm0.48$ & $23.82\pm0.36$ & $31.86\pm0.34$ & $65.76\pm0.22$\\
    Quadruplet (D) & $76.34\pm0.27$ & $84.67\pm0.23$ & $34.28\pm0.88$ & $23.49\pm0.32$ & $31.30\pm0.48$ & $64.79\pm0.50$\\
    SNR (D) & $78.69\pm0.19$ & $86.44\pm0.22$ & $35.88\pm0.71$ & $24.20\pm0.43$ & $32.17\pm0.60$ & $65.84\pm0.52$\\
    SoftTriple & $75.66\pm0.46$ & $83.72\pm0.29$ & $31.07\pm0.56$ & $22.90\pm0.25$ & $30.42\pm0.31$ & $62.66\pm0.16$\\
    Softmax & $78.91\pm0.27$ & $86.66\pm0.23$ & $35.51\pm0.85$ & $22.81\pm0.14$ & $29.84\pm0.13$ & $66.35\pm0.30$\\
    \midrule
    Triplet (D) & $79.13\pm0.27$ & $86.74\pm0.17$ & $35.89\pm0.25$ & $24.38\pm0.18$ & $32.26\pm0.25$ & $65.90\pm0.18$\\
    Triplet (R) & $70.63\pm0.43$ & $80.43\pm0.26$ & $29.02\pm0.47$ & $19.48\pm0.20$ & $25.97\pm0.28$ & $61.09\pm0.27$\\
    Triplet (S) & $72.51\pm0.47$ & $81.53\pm0.29$ & $31.61\pm0.41$ & $21.63\pm0.26$ & $28.32\pm0.56$ & $62.84\pm0.41$\\
    Triplet (H) & $77.60\pm0.33$ & $85.63\pm0.27$ & $34.71\pm0.31$ & $23.84\pm0.13$ & $31.31\pm0.15$ & $65.37\pm0.26$\\
    \midrule
    \midrule
    R-Contrastive (D) & $81.06\pm0.41$ & $88.06\pm0.21$ & $37.72\pm0.84$ & $24.55\pm0.34$ & $32.53\pm0.43$ & $67.27\pm0.46$\\
    R-Margin (D, $\beta=0.6$) & \blue{$\mathbf{82.37\pm0.13}$} & \blue{$\mathbf{89.14\pm0.12}$} & \blue{$\mathbf{39.28\pm0.41}$} & \blue{$\mathbf{25.67\pm0.32}$} & \blue{$\mathbf{34.57\pm0.30}$} & $68.66\pm0.47$\\
    R-Margin (D, $\beta=1.2$) & $81.11\pm0.49$ & $88.20\pm0.22$ & $38.76\pm0.94$ & $24.17\pm0.50$ & $31.60\pm0.79$ & $67.72\pm0.79$\\
    R-SNR (D) & $80.38\pm0.35$ & $87.95\pm0.37$ & $38.62\pm0.47$ & $24.72\pm0.15$ & $33.02\pm0.23$ & $67.60\pm0.20$\\
    R-Triplet (D) & $81.17\pm0.11$ & $88.43\pm0.18$ & $38.72\pm0.31$ & $25.27\pm0.22$ & $33.18\pm0.48$ & $67.79\pm0.23$\\
    \bottomrule
    \end{tabular}
    \caption{Comparison of DML setups for CARS196. We report all relevant performance metrics. Training is done over 150 epochs.}
    \label{tab:eval_car}
 \end{table*}

  \begin{table*}[h]
 \small
   \setlength\tabcolsep{1.4pt}
   \centering
   %\begin{tabular}{l|c|ccc|c}
   \begin{tabular}{l|c|c|c|c|c|c}
     \toprule
     \multicolumn{7}{c}{Stanford Online Products\cite{lifted}} \\
     \midrule
     Approach & R@1 & R@2 & F1 & mAP@C & mAP@1000 & NMI\\
     \midrule
    Imagenet & 48.65 & 53.82 & 11.97 & 17.47 & 18.58 & 58.64\\
    \midrule         
    Angular & $73.22\pm0.07$ & $78.14\pm0.06$ & $34.20\pm0.07$ & $36.96\pm0.07$ & $40.76\pm0.08$ & $89.53\pm0.01$\\
    ArcFace & $77.71\pm0.15$ & $82.23\pm0.09$ & $37.15\pm0.13$ & $41.20\pm0.11$ & $45.76\pm0.14$ & $90.09\pm0.03$\\
    Contrastive (D) & $73.21\pm0.04$ & $77.87\pm0.04$ & $35.66\pm0.14$ & $37.43\pm0.05$ & $41.21\pm0.06$ & $89.78\pm0.02$\\
    GenLifted & $75.21\pm0.12$ & $80.25\pm0.04$ & $35.93\pm0.09$ & $39.03\pm0.09$ & $43.06\pm0.14$ & $89.84\pm0.01$\\
    Histogram & $71.30\pm0.10$ & $76.18\pm0.08$ & $31.58\pm0.14$ & $34.88\pm0.10$ & $38.55\pm0.13$ & $88.93\pm0.02$\\
    Multisimilarity & $77.99\pm0.09$ & $82.64\pm0.08$ & $36.75\pm0.18$ & $41.52\pm0.07$ & $46.23\pm0.08$ & $90.00\pm0.02$\\
    Margin (D, $\beta=0.6$) & $77.38\pm0.11$ & $81.78\pm0.12$ & \blue{$\mathbf{39.04\pm0.16}$} & $41.69\pm0.14$ & $45.92\pm0.17$ & \blue{$\mathbf{90.45\pm0.03}$}\\
    Margin (D, $\beta=1.2$) & $78.43\pm0.07$ & $82.83\pm0.09$ & $38.63\pm0.18$ & $42.43\pm0.12$ & $46.90\pm0.16$ & $90.40\pm0.03$\\
    Npair & $75.86\pm0.08$ & $80.73\pm0.06$ & $35.40\pm0.15$ & $39.09\pm0.10$ & $43.22\pm0.11$ & $89.79\pm0.03$\\
    Quadruplet (D) & $76.95\pm0.10$ & $81.54\pm0.05$ & $37.43\pm0.14$ & $40.82\pm0.13$ & $45.04\pm0.12$ & $90.14\pm0.02$\\
    SNR (D) & $77.61\pm0.34$ & $82.34\pm0.31$ & $37.17\pm0.37$ & $41.47\pm0.38$ & $46.03\pm0.47$ & $90.10\pm0.08$\\
    Softmax & $76.92\pm0.64$ & $81.34\pm0.63$ & $36.01\pm0.71$ & $40.23\pm0.78$ & $44.29\pm0.78$ & $89.82\pm0.15$\\
    \midrule
    Triplet (D) & $77.39\pm0.15$ & $82.03\pm0.08$ & $36.98\pm0.11$ & $41.02\pm0.12$ & $45.64\pm0.14$ & $90.06\pm0.02$\\
    Triplet (R) & $67.86\pm0.14$ & $73.02\pm0.11$ & $28.98\pm0.12$ & $31.92\pm0.17$ & $35.46\pm0.20$ & $88.35\pm0.04$\\
    Triplet (S) & $73.61\pm0.14$ & $78.36\pm0.14$ & $33.65\pm0.13$ & $37.11\pm0.06$ & $41.24\pm0.06$ & $89.35\pm0.02$\\
    Triplet (H) & $73.50\pm0.09$ & $78.38\pm0.06$ & $33.01\pm0.20$ & $36.85\pm0.05$ & $40.64\pm0.07$ & $89.25\pm0.03$\\
    \midrule
    \midrule
    R-Contrastive (D) & $74.36\pm0.11$ & $78.85\pm0.11$ & $36.39\pm0.07$ & $38.45\pm0.14$ & $42.60\pm0.14$ & $89.94\pm0.02$\\
    R-Margin (D, $\beta=0.6$) & $77.58\pm0.11$ & $81.93\pm0.10$ & $38.87\pm0.19$ & $41.74\pm0.12$ & $46.02\pm0.12$ & $90.42\pm0.03$\\
    R-Margin (D, $\beta=1.2$) & \blue{$\mathbf{78.52\pm0.10}$} & \blue{$\mathbf{82.95\pm0.07}$} & $38.36\pm0.16$ & \blue{$\mathbf{42.55\pm0.10}$} & \blue{$\mathbf{46.77\pm0.11}$} & $90.33\pm0.02$\\
    R-SNR (D) & $77.69\pm0.25$ & $82.46\pm0.17$ & $36.78\pm0.36$ & $41.44\pm0.31$ & $45.74\pm0.38$ & $90.02\pm0.06$\\
    R-Triplet (D) & $77.33\pm0.14$ & $82.01\pm0.12$ & $36.63\pm0.20$ & $40.91\pm0.12$ & $45.57\pm0.14$ & $89.98\pm0.04$\\

    \bottomrule
    \end{tabular}
    \caption{Comparison of DML setups for Stanford Online Products. We report all relevant performance metrics.. Training is done over 100 epochs.}
    \label{tab:eval_sop}
 \end{table*}

%% file: tables/total_comparison.tex
 \begin{table*}[t]
 \small
   \setlength\tabcolsep{1.4pt}
   \centering
   %\begin{tabular}{l|c|ccc|c}
   \begin{tabular}{l|c|c|c|c|c|}
     \toprule
     \multicolumn{6}{c}{CUB200-2011\cite{cub200-2011}} \\
     \midrule
     Approach & R@1 & R@2 & F1 & mAP@C & NMI\\
     \midrule
    Histogram, SPC-2 & $57.92\pm0.26$ & $69.74\pm0.03$ & $31.21\pm0.18$ & $21.27\pm0.28$ & $63.26\pm0.18$\\
    Histogram, SPC-4 & $57.88\pm0.35$ & $70.18\pm0.28$ & $31.15\pm0.17$ & $21.39\pm0.26$ & $63.37\pm0.12$\\
    Histogram, SPC-8 & $57.87\pm0.40$ & $70.09\pm0.41$ & $31.72\pm0.09$ & $21.56\pm0.07$ & $63.51\pm0.13$\\
    Histogram, DDM   & $58.22\pm0.33$ & $70.61\pm0.26$ & $32.08\pm0.21$ & $21.69\pm0.26$ & $63.55\pm0.20$\\
    Histogram, GC    & $57.51\pm0.40$ & $69.64\pm0.25$ & $30.98\pm0.43$ & $21.16\pm0.25$ & $63.08\pm0.18$\\
    Histogram, SPC-R & $57.62\pm0.11$ & $69.37\pm0.22$ & $30.82\pm0.29$ & $21.03\pm0.12$ & $63.03\pm0.28$\\
    Histogram, FRD   & $58.36\pm0.19$ & $70.79\pm0.31$ & $32.12\pm0.29$ & $21.67\pm0.21$ & $63.81\pm0.24$\\
    Margin (D), SPC-2 & $62.66\pm0.28$ & $73.98\pm0.08$ & $37.77\pm0.51$ & $23.40\pm0.22$ & $67.76\pm0.19$\\
    Margin (D), SPC-4 & $62.37\pm0.25$ & $74.05\pm0.23$ & $37.84\pm0.30$ & $23.34\pm0.16$ & $67.90\pm0.15$\\
    Margin (D), SPC-8 & $62.04\pm0.12$ & $73.87\pm0.22$ & $37.03\pm0.44$ & $23.21\pm0.29$ & $67.36\pm0.20$\\
    Margin (D), DDM   & $62.50\pm0.23$ & $74.31\pm0.24$ & $37.90\pm0.41$ & $23.32\pm0.19$ & $68.00\pm0.25$\\
    Margin (D), GC    & $62.61\pm0.26$ & $74.29\pm0.27$ & $37.84\pm0.83$ & $23.43\pm0.23$ & $67.81\pm0.46$\\
    Margin (D), SPC-R & $62.36\pm0.39$ & $74.14\pm0.36$ & $37.62\pm0.55$ & $23.23\pm0.28$ & $67.47\pm0.25$\\
    Margin (D), FRD   & $62.64\pm0.34$ & $74.08\pm0.34$ & $38.11\pm0.69$ & $23.37\pm0.21$ & $67.87\pm0.33$\\
    MultiSimilarity, SPC-2 & $62.46\pm0.25$ & $74.13\pm0.13$ & $38.61\pm0.42$ & $22.26\pm0.14$ & $68.00\pm0.20$\\
    MultiSimilarity, SPC-4 & $62.95\pm0.12$ & $74.61\pm0.02$ & $38.39\pm0.37$ & $22.66\pm0.12$ & $68.29\pm0.22$\\
    MultiSimilarity, SPC-8 & $62.73\pm0.19$ & $74.22\pm0.05$ & $37.18\pm0.07$ & $22.82\pm0.12$ & $67.46\pm0.07$\\
    MultiSimilarity, DDM & $62.57\pm0.31$ & $74.49\pm0.25$ & $38.58\pm0.70$ & $22.31\pm0.16$ & $68.25\pm0.32$\\
    MultiSimilarity, GC & $62.65\pm0.24$ & $74.21\pm0.28$ & $38.79\pm0.20$ & $22.35\pm0.11$ & $68.08\pm0.29$\\
    MultiSimilarity, SPC-R & $62.36\pm0.32$ & $74.10\pm0.33$ & $37.99\pm0.39$ & $22.32\pm0.22$ & $68.01\pm0.20$\\
    MultiSimilarity, FRD & $63.19\pm0.31$ & $74.88\pm0.27$ & $38.70\pm0.57$ & $23.02\pm0.22$ & $68.24\pm0.26$\\
    NPair, SPC-2 & $60.52\pm0.88$ & $73.12\pm0.86$ & $36.51\pm0.55$ & $22.12\pm0.23$ & $66.79\pm0.55$\\
    NPair, SPC-4 & $59.80\pm0.20$ & $71.42\pm0.29$ & $34.23\pm0.55$ & $21.59\pm0.17$ & $65.31\pm0.41$\\
    NPair, SPC-8 & $58.22\pm0.07$ & $70.29\pm0.02$ & $33.07\pm0.38$ & $20.84\pm0.09$ & $64.29\pm0.11$\\
    NPair, DDM & $60.13\pm0.05$ & $72.03\pm0.15$ & $35.24\pm0.42$ & $21.69\pm0.01$ & $66.05\pm0.32$\\
    NPair, GC & $60.85\pm0.44$ & $72.90\pm0.52$ & $36.13\pm0.68$ & $22.12\pm0.16$ & $66.71\pm0.40$\\
    NPair, SPC-R & $61.32\pm0.07$ & $73.08\pm0.26$ & $36.45\pm0.19$ & $22.28\pm0.17$ & $66.87\pm0.25$\\
    NPair, FRD   & $61.23\pm0.15$ & $73.01\pm0.17$ & $36.26\pm0.26$ & $22.34\pm0.17$ & $67.04\pm0.22$\\
    ProxyNCA, SPC-2 & $62.67\pm0.43$ & $73.96\pm0.36$ & $35.66\pm0.26$ & $23.64\pm0.52$ & $66.88\pm0.29$\\
    ProxyNCA, SPC-4 & $62.50\pm0.48$ & $73.64\pm0.47$ & $35.46\pm0.62$ & $23.50\pm0.44$ & $66.59\pm0.32$\\
    ProxyNCA, SPC-8 & $62.49\pm0.39$ & $74.07\pm0.31$ & $35.44\pm0.60$ & $23.90\pm0.54$ & $66.56\pm0.32$\\
    ProxyNCA, DDM & $62.63\pm0.00$ & $73.68\pm0.00$ & $36.35\pm0.00$ & $24.50\pm0.00$ & $67.08\pm0.00$\\
    ProxyNCA, GC & $62.97\pm0.53$ & $74.03\pm0.50$ & $36.67\pm0.96$ & $24.17\pm0.43$ & $67.15\pm0.51$\\
    ProxyNCA, SPC-R & $62.99\pm0.84$ & $74.07\pm0.42$ & $36.61\pm0.77$ & $23.96\pm0.39$ & $67.26\pm0.78$\\
    ProxyNCA, FRD & $63.12\pm0.51$ & $74.36\pm0.31$ & $37.37\pm0.58$ & $24.42\pm0.30$ & $67.54\pm0.46$\\
    Softmax, SPC-2 & $61.51\pm0.28$ & $73.29\pm0.23$ & $35.36\pm0.55$ & $22.02\pm0.07$ & $66.43\pm0.30$\\
    Softmax, SPC-4 & $61.55\pm0.50$ & $73.51\pm0.22$ & $35.72\pm0.12$ & $22.08\pm0.15$ & $66.63\pm0.20$\\
    Softmax, SPC-8 & $61.55\pm0.49$ & $73.29\pm0.24$ & $35.35\pm0.34$ & $22.16\pm0.03$ & $66.22\pm0.30$\\
    Softmax, DDM & $61.72\pm0.78$ & $72.99\pm0.30$ & $36.65\pm1.02$ & $22.67\pm0.35$ & $66.79\pm0.29$\\
    Softmax, GC & $61.32\pm0.43$ & $72.84\pm0.29$ & $36.58\pm0.46$ & $22.51\pm0.32$ & $66.83\pm0.11$\\
    Softmax, SPC-R & $61.58\pm0.23$ & $73.30\pm0.23$ & $35.38\pm0.49$ & $21.89\pm0.13$ & $66.46\pm0.32$\\
    Softmax, FRD & $61.52\pm0.69$ & $72.75\pm0.63$ & $35.98\pm0.89$ & $22.23\pm0.37$ & $66.48\pm0.36$\\
    Triplet (R), SPC-2 & $58.44\pm0.89$ & $70.42\pm0.41$ & $31.94\pm0.57$ & $20.72\pm0.12$ & $63.98\pm0.22$\\
    Triplet (R), SPC-4 & $58.67\pm0.43$ & $70.79\pm0.32$ & $32.18\pm0.45$ & $20.86\pm0.25$ & $64.24\pm0.35$\\
    Triplet (R), SPC-8 & $58.04\pm0.23$ & $70.22\pm0.29$ & $32.26\pm0.18$ & $20.67\pm0.15$ & $63.74\pm0.09$\\
    Triplet (R), DDM & $58.08\pm0.45$ & $70.00\pm0.15$ & $31.58\pm0.33$ & $20.65\pm0.07$ & $63.56\pm0.08$\\
    Triplet (R), GC & $58.42\pm0.13$ & $70.26\pm0.13$ & $31.78\pm0.17$ & $20.70\pm0.07$ & $63.96\pm0.13$\\
    Triplet (R), SPC-R & $58.33\pm0.38$ & $70.50\pm0.28$ & $31.32\pm0.23$ & $20.91\pm0.10$ & $63.50\pm0.17$\\
    Triplet (R), FRD & $58.00\pm0.00$ & $69.95\pm0.15$ & $31.42\pm0.11$ & $20.33\pm0.19$ & $63.46\pm0.01$\\
    \bottomrule
   \end{tabular}
     \caption{CUB200-2011: Comparison of Batch-Sampling methods for various loss functions and sampling methods.}
     \label{tab:meta_sampling_cub}
 \end{table*}

 \begin{table*}[t]
 \small
   \setlength\tabcolsep{1.4pt}
   \centering
   %\begin{tabular}{l|c|ccc|c}
   \begin{tabular}{l|c|c|c|c|c|}
     \toprule
     \multicolumn{6}{c}{CARS\cite{cars196}} \\
     \midrule
     Approach & R@1 & R@2 & F1 & mAP@C & NMI\\
     \midrule
    Histogram, SPC-2 & $67.24\pm1.04$ & $77.37\pm0.85$ & $28.93\pm0.65$ & $19.89\pm0.49$ & $60.82\pm0.59$\\
    Histogram, SPC-4 & $67.40\pm0.53$ & $77.53\pm0.43$ & $28.54\pm0.74$ & $19.85\pm0.24$ & $60.97\pm0.58$\\
    Histogram, SPC-8 & $67.53\pm0.77$ & $77.69\pm0.56$ & $29.14\pm0.76$ & $20.04\pm0.35$ & $61.22\pm0.38$\\
    Histogram, DDM  & $66.57\pm0.49$ & $76.93\pm0.53$ & $27.99\pm0.41$ & $19.55\pm0.28$ & $60.46\pm0.38$\\
    Histogram, GC  & $66.36\pm0.76$ & $76.88\pm0.57$ & $27.70\pm0.55$ & $19.04\pm0.47$ & $60.40\pm0.45$\\
    Histogram, SPC-R & $64.34\pm0.65$ & $75.43\pm0.49$ & $27.07\pm0.75$ & $18.37\pm0.26$ & $59.90\pm0.75$\\
    Histogram, FRD & $67.06\pm0.18$ & $77.18\pm0.22$ & $28.19\pm0.81$ & $19.65\pm0.22$ & $60.31\pm0.29$\\
    Margin (D), SPC-2 & $79.79\pm0.40$ & $87.27\pm0.36$ & $38.78\pm0.48$ & $24.84\pm0.28$ & $67.59\pm0.24$\\
    Margin (D), SPC-4 & $79.73\pm0.08$ & $87.18\pm0.11$ & $38.29\pm0.41$ & $25.13\pm0.21$ & $67.48\pm0.46$\\
    Margin (D), SPC-8 & $78.93\pm0.23$ & $86.71\pm0.18$ & $37.18\pm0.29$ & $24.51\pm0.37$ & $66.83\pm0.24$\\
    Margin (D), DDM & $80.13\pm0.38$ & $87.53\pm0.12$ & $38.26\pm0.24$ & $24.65\pm0.12$ & $67.32\pm0.26$\\
    Margin (D), GC & $80.22\pm0.16$ & $87.44\pm0.03$ & $37.91\pm0.71$ & $24.82\pm0.34$ & $67.14\pm0.39$\\
    Margin (D), SPC-R & $80.06\pm0.48$ & $87.37\pm0.30$ & $38.17\pm1.01$ & $24.54\pm0.21$ & $67.26\pm0.37$\\
    Margin (D), FRD & $80.23\pm0.20$ & $87.73\pm0.10$ & $38.59\pm0.59$ & $25.18\pm0.12$ & $67.56\pm0.21$\\
    MultiSimilarity, SPC-2 & $81.59\pm0.18$ & $88.92\pm0.07$ & $40.79\pm0.69$ & $24.35\pm0.25$ & $69.63\pm0.50$\\
    MultiSimilarity, SPC-4 & $81.78\pm0.13$ & $88.97\pm0.13$ & $41.02\pm0.23$ & $25.15\pm0.16$ & $69.47\pm0.12$\\
    MultiSimilarity, SPC-8 & $81.32\pm0.05$ & $88.28\pm0.10$ & $39.09\pm0.71$ & $25.54\pm0.23$ & $68.36\pm0.42$\\
    MultiSimilarity, DDM & $81.77\pm0.29$ & $88.77\pm0.24$ & $40.94\pm0.32$ & $23.80\pm0.06$ & $69.26\pm0.24$\\
    MultiSimilarity, GC & $81.63\pm0.11$ & $88.72\pm0.22$ & $40.38\pm0.21$ & $24.27\pm0.27$ & $69.36\pm0.22$\\
    MultiSimilarity, SPC-R & $81.52\pm0.16$ & $88.74\pm0.16$ & $40.66\pm0.31$ & $24.67\pm0.17$ & $69.63\pm0.22$\\
    MultiSimilarity, FRD & $81.70\pm0.18$ & $88.96\pm0.19$ & $41.74\pm0.39$ & $24.25\pm0.10$ & $69.56\pm0.17$\\
    NPair, SPC-2 & $76.35\pm0.23$ & $84.79\pm0.17$ & $35.72\pm0.49$ & $23.45\pm0.10$ & $66.22\pm0.09$\\
    NPair, SPC-4 & $73.57\pm0.15$ & $82.76\pm0.20$ & $33.94\pm0.21$ & $22.83\pm0.15$ & $64.96\pm0.18$\\
    NPair, SPC-8 & $71.97\pm0.35$ & $81.98\pm0.30$ & $32.69\pm0.40$ & $22.57\pm0.15$ & $63.99\pm0.22$\\
    NPair, DDM & $76.02\pm0.32$ & $84.47\pm0.03$ & $35.35\pm0.07$ & $23.40\pm0.03$ & $66.11\pm0.11$\\
    NPair, GC & $76.09\pm0.11$ & $84.64\pm0.24$ & $35.03\pm0.26$ & $23.23\pm0.19$ & $65.83\pm0.15$\\
    NPair, SPC-R & $75.79\pm0.09$ & $84.64\pm0.12$ & $35.14\pm0.44$ & $23.07\pm0.13$ & $65.91\pm0.22$\\
    NPair, FRD & $75.83\pm0.49$ & $84.49\pm0.25$ & $35.65\pm0.70$ & $23.32\pm0.36$ & $66.08\pm0.53$\\
    ProxyNCA, SPC-2 & $78.48\pm0.61$ & $85.97\pm0.39$ & $34.82\pm0.58$ & $23.85\pm0.38$ & $65.74\pm0.15$\\
    ProxyNCA, SPC-4 & $78.48\pm0.61$ & $85.94\pm0.25$ & $34.90\pm0.57$ & $23.77\pm0.20$ & $65.55\pm0.44$\\
    ProxyNCA, SPC-8 & $78.08\pm0.20$ & $85.84\pm0.28$ & $33.35\pm1.17$ & $23.30\pm0.25$ & $65.26\pm0.62$\\
    ProxyNCA, DDM & $78.43\pm0.30$ & $86.30\pm0.26$ & $34.72\pm0.65$ & $23.62\pm0.20$ & $65.84\pm0.32$\\
    ProxyNCA, GC & $78.14\pm0.55$ & $85.92\pm0.42$ & $34.72\pm0.34$ & $23.43\pm0.23$ & $65.60\pm0.28$\\
    ProxyNCA, SPC-R & $78.45\pm0.23$ & $86.19\pm0.21$ & $35.18\pm0.75$ & $23.91\pm0.19$ & $66.19\pm0.39$\\
    ProxyNCA, FRD & $78.43\pm0.06$ & $87.09\pm0.15$ & $34.78\pm0.43$ & $23.72\pm0.08$ & $65.70\pm0.15$\\
    Softmax, SPC-2 & $79.76\pm0.26$ & $87.70\pm0.30$ & $35.94\pm0.33$ & $24.04\pm0.29$ & $67.57\pm0.27$\\
    Softmax, SPC-4 & $79.42\pm0.39$ & $87.47\pm0.20$ & $35.80\pm0.59$ & $23.91\pm0.30$ & $67.30\pm0.37$\\
    Softmax, SPC-8 & $79.53\pm0.38$ & $87.30\pm0.22$ & $35.22\pm0.64$ & $24.03\pm0.22$ & $67.03\pm0.29$\\
    Softmax, DDM & $79.23\pm0.32$ & $87.40\pm0.29$ & $35.38\pm0.59$ & $23.52\pm0.21$ & $67.02\pm0.27$\\
    Softmax, GC & $79.22\pm0.36$ & $87.38\pm0.36$ & $35.50\pm0.59$ & $23.55\pm0.25$ & $67.24\pm0.15$\\
    Softmax, SPC-R & $79.53\pm0.18$ & $87.71\pm0.09$ & $36.71\pm0.23$ & $24.12\pm0.26$ & $67.79\pm0.39$\\
    Softmax, FRD & $79.25\pm0.41$ & $87.49\pm0.48$ & $35.36\pm0.21$ & $23.47\pm0.26$ & $67.12\pm0.18$\\
    Triplet (R), SPC-2 & $69.73\pm0.47$ & $79.74\pm0.28$ & $28.58\pm0.28$ & $19.03\pm0.19$ & $60.64\pm0.38$\\
    Triplet (R), SPC-4 & $69.86\pm0.49$ & $79.91\pm0.51$ & $28.84\pm0.18$ & $19.11\pm0.08$ & $60.97\pm0.16$\\
    Triplet (R), SPC-8 & $69.32\pm0.23$ & $79.43\pm0.64$ & $28.38\pm0.11$ & $19.09\pm0.03$ & $60.63\pm0.17$\\
    Triplet (R), DDM & $69.78\pm0.25$ & $79.87\pm0.35$ & $28.07\pm0.41$ & $18.78\pm0.32$ & $60.38\pm0.24$\\
    Triplet (R), GC & $69.34\pm0.29$ & $79.41\pm0.18$ & $28.68\pm0.78$ & $18.89\pm0.30$ & $60.87\pm0.36$\\
    Triplet (R), SPC-R & $69.01\pm0.38$ & $79.33\pm0.16$ & $27.90\pm0.28$ & $18.43\pm0.33$ & $60.43\pm0.26$\\
    Triplet (R), FRD & $69.55\pm0.58$ & $79.52\pm0.54$ & $28.70\pm0.75$ & $18.77\pm0.35$ & $60.83\pm0.51$\\
    \bottomrule
   \end{tabular}
     \caption{CARS196: Comparison of Batch-Sampling methods for various loss functions and sampling methods.}
     \label{tab:meta_sampling_cars}
 \end{table*}

 \begin{table*}[t]
 \small
   \setlength\tabcolsep{1.4pt}
   \centering
   %\begin{tabular}{l|c|ccc|c}
   \begin{tabular}{l|c|c|c|c|c|}
     \toprule
     \multicolumn{6}{c}{Stanford Online-Products\cite{lifted}} \\
     \midrule
     Approach & R@1 & R@2 & F1 & mAP & NMI\\
     \midrule
    Histogram, SPC-2 & $69.52\pm0.10$ & $74.57\pm0.09$ & $30.49\pm0.05$ & $33.20\pm0.07$ & $88.69\pm0.01$\\
    Histogram, SPC-4 & $70.13\pm0.24$ & $75.17\pm0.21$ & $31.05\pm0.00$ & $33.85\pm0.13$ & $88.82\pm0.01$\\
    Histogram, DDM & $69.36\pm0.06$ & $74.39\pm0.09$ & $30.47\pm0.13$ & $33.20\pm0.03$ & $88.69\pm0.01$\\
    Histogram, GC & $68.42\pm0.22$ & $73.66\pm0.15$ & $30.03\pm0.25$ & $32.49\pm0.24$ & $88.58\pm0.05$\\
    Histogram, SPC-R & $59.06\pm0.19$ & $64.72\pm0.06$ & $22.74\pm0.14$ & $25.16\pm0.07$ & $86.90\pm0.03$\\
    Histogram, FRD & $69.71\pm0.19$ & $74.84\pm0.17$ & $30.65\pm0.21$ & $33.45\pm0.07$ & $88.71\pm0.05$\\
    Margin (D), SPC-2 & $78.28\pm0.08$ & $82.69\pm0.03$ & $38.28\pm0.11$ & $42.36\pm0.11$ & $90.34\pm0.03$\\
    Margin (D), SPC-4 & $77.51\pm0.14$ & $81.91\pm0.13$ & $37.52\pm0.34$ & $41.41\pm0.22$ & $90.19\pm0.06$\\
    Margin (D), DDM & $77.90\pm0.13$ & $82.28\pm0.23$ & $37.61\pm0.68$ & $41.82\pm0.26$ & $90.22\pm0.11$\\
    Margin (D), GC & $75.77\pm0.40$ & $80.40\pm0.41$ & $35.45\pm0.63$ & $39.55\pm0.43$ & $89.72\pm0.13$\\
    Margin (D), SPC-R & $68.28\pm0.08$ & $73.37\pm0.09$ & $25.64\pm0.23$ & $31.31\pm0.13$ & $87.55\pm0.05$\\
    Margin (D), FRD & $78.18\pm0.18$ & $82.60\pm0.18$ & $38.25\pm0.53$ & $42.20\pm0.31$ & $90.34\pm0.20$\\
    MultiSimilarity, SPC-2 & $77.80\pm0.07$ & $82.47\pm0.06$ & $36.37\pm0.08$ & $41.31\pm0.02$ & $89.93\pm0.03$\\
    MultiSimilarity, SPC-4 & $77.90\pm0.13$ & $82.53\pm0.04$ & $36.98\pm0.10$ & $41.51\pm0.05$ & $90.06\pm0.06$\\
    MultiSimilarity, DDM & $77.85\pm0.03$ & $82.60\pm0.05$ & $36.57\pm0.22$ & $41.35\pm0.12$ & $89.96\pm0.06$\\
    MultiSimilarity, GC & $76.51\pm0.23$ & $81.28\pm0.17$ & $35.24\pm0.28$ & $39.92\pm0.27$ & $89.67\pm0.06$\\
    MultiSimilarity, SPC-R & $72.16\pm0.38$ & $76.12\pm0.23$ & $31.77\pm0.17$ & $35.01\pm0.17$ & $88.25\pm0.05$\\
    MultiSimilarity, FRD & $77.97\pm0.17$ & $82.60\pm0.22$ & $36.44\pm0.31$ & $41.54\pm0.27$ & $89.95\pm0.03$\\
    NPair, SPC-2 & $75.42\pm0.16$ & $80.36\pm0.14$ & $34.60\pm0.29$ & $38.49\pm0.27$ & $89.63\pm0.06$\\
    NPair, SPC-4 & $70.42\pm0.24$ & $75.90\pm0.25$ & $32.60\pm0.37$ & $34.81\pm0.21$ & $89.11\pm0.06$\\
    NPair, DDM & $74.12\pm0.32$ & $79.20\pm0.29$ & $33.39\pm0.54$ & $36.99\pm0.35$ & $89.42\pm0.09$\\
    NPair, GC & $74.37\pm0.31$ & $79.27\pm0.36$ & $33.55\pm0.67$ & $37.47\pm0.49$ & $89.39\pm0.14$\\
    NPair, SPC-R & $68.23\pm0.04$ & $73.45\pm0.04$ & $28.26\pm0.15$ & $31.80\pm0.02$ & $88.19\pm0.02$\\
    NPair, FRD & $75.50\pm0.41$ & $80.43\pm0.43$ & $35.36\pm0.52$ & $38.98\pm0.30$ & $89.77\pm0.13$\\
    Softmax, SPC-2 & $78.12\pm0.21$ & $82.39\pm0.16$ & $38.12\pm0.21$ & $42.17\pm0.13$ & $90.00\pm0.09$\\
    Softmax, SPC-4 & $77.81\pm0.28$ & $82.21\pm0.19$ & $38.16\pm0.17$ & $42.12\pm0.14$ & $90.08\pm0.02$\\
    Softmax, DDM   & $77.39\pm0.33$ & $81.67\pm0.26$ & $37.64\pm0.26$ & $41.97\pm0.20$ & $89.66\pm0.10$\\
    Softmax, GC    & $78.50\pm0.29$ & $82.47\pm0.24$ & $38.68\pm0.27$ & $42.37\pm0.31$ & $90.22\pm0.16$\\
    Softmax, SPC-R & $78.38\pm0.19$ & $82.46\pm0.10$ & $38.64\pm0.13$ & $42.29\pm0.30$ & $90.31\pm0.16$\\
    Softmax, FRD   & $77.58\pm0.22$ & $81.93\pm0.23$ & $38.01\pm0.43$ & $42.23\pm0.19$ & $90.08\pm0.06$\\
    Triplet (R), SPC-2 & $66.86\pm0.36$ & $72.06\pm0.28$ & $27.38\pm0.29$ & $30.78\pm0.23$ & $88.01\pm0.09$\\
    Triplet (R), SPC-4 & $67.13\pm0.45$ & $72.29\pm0.39$ & $27.46\pm0.15$ & $30.99\pm0.18$ & $88.02\pm0.04$\\
    Triplet (R), DDM & $66.96\pm0.11$ & $72.18\pm0.10$ & $27.47\pm0.02$ & $30.79\pm0.01$ & $88.01\pm0.01$\\
    Triplet (R), GC & $66.61\pm0.14$ & $71.75\pm0.03$ & $27.49\pm0.22$ & $30.44\pm0.04$ & $87.99\pm0.03$\\
    Triplet (R), SPC-R & $61.12\pm0.02$ & $66.39\pm0.04$ & $23.02\pm0.09$ & $26.07\pm0.15$ & $86.95\pm0.01$\\
    Triplet (R), FRD & $67.00\pm0.22$ & $72.04\pm0.15$ & $27.32\pm0.16$ & $30.79\pm0.20$ & $87.95\pm0.10$\\
    \bottomrule
   \end{tabular}
     \caption{SOP: Comparison of Batch-Sampling methods for various loss functions and sampling methods.}
     \label{tab:meta_sampling_sop}
 \end{table*}

%% file: main.bbl
\begin{thebibliography}{71}
\providecommand{\natexlab}[1]{#1}
\providecommand{\url}[1]{\texttt{#1}}
\expandafter\ifx\csname urlstyle\endcsname\relax
  \providecommand{\doi}[1]{doi: #1}\else
  \providecommand{\doi}{doi: \begingroup \urlstyle{rm}\Url}\fi

\bibitem[Achille \& Soatto(2016)Achille and Soatto]{alex2016information}
Achille, A. and Soatto, S.
\newblock Information dropout: Learning optimal representations through noisy
  computation, 2016.

\bibitem[Agarwal et~al.(2005)Agarwal, Har-Peled, and Varadarajan]{coresets}
Agarwal, P.~K., Har-Peled, S., and Varadarajan, K.~R.
\newblock Geometric approximation via coresets.
\newblock \emph{Combinatorial and computational geometry}, 52:\penalty0 1--30,
  2005.

\bibitem[Alemi et~al.(2016)Alemi, Fischer, Dillon, and Murphy]{alex2016deep}
Alemi, A.~A., Fischer, I., Dillon, J.~V., and Murphy, K.
\newblock Deep variational information bottleneck, 2016.

\bibitem[Belghazi et~al.(2018)Belghazi, Baratin, Rajeswar, Ozair, Bengio,
  Courville, and Hjelm]{belghazi2018mutual}
Belghazi, M.~I., Baratin, A., Rajeswar, S., Ozair, S., Bengio, Y., Courville,
  A., and Hjelm, R.~D.
\newblock Mine: Mutual information neural estimation, 2018.

\bibitem[Bellet(2013)]{bellet2013supervised}
Bellet, A.
\newblock Supervised metric learning with generalization guarantees, 2013.

\bibitem[Bellet \& Habrard(2015)Bellet and Habrard]{Bellet_2015}
Bellet, A. and Habrard, A.
\newblock Robustness and generalization for metric learning.
\newblock \emph{Neurocomputing}, 151:\penalty0 259–267, Mar 2015.
\newblock ISSN 0925-2312.
\newblock \doi{10.1016/j.neucom.2014.09.044}.
\newblock URL \url{http://dx.doi.org/10.1016/j.neucom.2014.09.044}.

\bibitem[Bouchacourt et~al.(2018)Bouchacourt, Tomioka, and Nowozin]{grouping}
Bouchacourt, D., Tomioka, R., and Nowozin, S.
\newblock Multi-level variational autoencoder: Learning disentangled
  representations from grouped observations.
\newblock In \emph{AAAI 2018}, 2018.

\bibitem[Bouthillier et~al.(2019)Bouthillier, Laurent, and
  Vincent]{bouthillier2019unreproducible}
Bouthillier, X., Laurent, C., and Vincent, P.
\newblock Unreproducible research is reproducible.
\newblock In \emph{International Conference on Machine Learning}, pp.\
  725--734, 2019.

\bibitem[Brock et~al.(2018)Brock, Donahue, and Simonyan]{BigGAN}
Brock, A., Donahue, J., and Simonyan, K.
\newblock Large scale {GAN} training for high fidelity natural image synthesis.
\newblock \emph{CoRR}, abs/1809.11096, 2018.
\newblock URL \url{http://arxiv.org/abs/1809.11096}.

\bibitem[Cakir et~al.(2019)Cakir, He, Xia, Kulis, and Sclaroff]{learn2rank}
Cakir, F., He, K., Xia, X., Kulis, B., and Sclaroff, S.
\newblock Deep metric learning to rank.
\newblock In \emph{The IEEE Conference on Computer Vision and Pattern
  Recognition (CVPR)}, June 2019.

\bibitem[Chen et~al.(2017)Chen, Chen, Zhang, and Huang]{quadtruplet}
Chen, W., Chen, X., Zhang, J., and Huang, K.
\newblock Beyond triplet loss: a deep quadruplet network for person
  re-identification.
\newblock In \emph{Proceedings of the IEEE Conference on Computer Vision and
  Pattern Recognition}, 2017.

\bibitem[Deng et~al.(2009)Deng, Dong, Socher, Li, Li, and Fei-Fei]{imagenet}
Deng, J., Dong, W., Socher, R., Li, L.-J., Li, K., and Fei-Fei, L.
\newblock {ImageNet: A Large-Scale Hierarchical Image Database}.
\newblock In \emph{IEEE Conference on Computer Vision and Pattern Recognition
  (CVPR)}, 2009.

\bibitem[Deng et~al.(2018)Deng, Guo, Xue, and Zafeiriou]{arcface}
Deng, J., Guo, J., Xue, N., and Zafeiriou, S.
\newblock Arcface: Additive angular margin loss for deep face recognition,
  2018.

\bibitem[Ge(2018)]{htl}
Ge, W.
\newblock Deep metric learning with hierarchical triplet loss.
\newblock In \emph{Proceedings of the European Conference on Computer Vision
  (ECCV)}, pp.\  269--285, 2018.

\bibitem[Goyal et~al.(2019)Goyal, Islam, Strouse, Ahmed, Botvinick, Larochelle,
  Bengio, and Levine]{goyal2019infobot}
Goyal, A., Islam, R., Strouse, D., Ahmed, Z., Botvinick, M., Larochelle, H.,
  Bengio, Y., and Levine, S.
\newblock Infobot: Transfer and exploration via the information bottleneck,
  2019.

\bibitem[Goyal et~al.(2017)Goyal, Doll{\'a}r, Girshick, Noordhuis, Wesolowski,
  Kyrola, Tulloch, Jia, and He]{goyal2017accurate}
Goyal, P., Doll{\'a}r, P., Girshick, R., Noordhuis, P., Wesolowski, L., Kyrola,
  A., Tulloch, A., Jia, Y., and He, K.
\newblock Accurate, large minibatch sgd: Training imagenet in 1 hour.
\newblock \emph{arXiv preprint arXiv:1706.02677}, 2017.

\bibitem[Hadsell et~al.(2006)Hadsell, Chopra, and LeCun]{contrastive}
Hadsell, R., Chopra, S., and LeCun, Y.
\newblock Dimensionality reduction by learning an invariant mapping.
\newblock In \emph{Proceedings of the IEEE Conference on Computer Vision and
  Pattern Recognition}, 2006.

\bibitem[Harwood et~al.(2017)Harwood, Kumar, Carneiro, Reid, Drummond,
  et~al.]{smartmining}
Harwood, B., Kumar, B., Carneiro, G., Reid, I., Drummond, T., et~al.
\newblock Smart mining for deep metric learning.
\newblock In \emph{Proceedings of the IEEE International Conference on Computer
  Vision}, pp.\  2821--2829, 2017.

\bibitem[He et~al.(2016)He, Zhang, Ren, and Sun]{resnet}
He, K., Zhang, X., Ren, S., and Sun, J.
\newblock Deep residual learning for image recognition.
\newblock In \emph{Proceedings of the IEEE conference on computer vision and
  pattern recognition}, pp.\  770--778, 2016.

\bibitem[Hermans et~al.(2017)Hermans, Beyer, and Leibe]{genlifted}
Hermans, A., Beyer, L., and Leibe, B.
\newblock In defense of the triplet loss for person re-identification, 2017.

\bibitem[Heusel et~al.(2017)Heusel, Ramsauer, Unterthiner, Nessler, and
  Hochreiter]{fid}
Heusel, M., Ramsauer, H., Unterthiner, T., Nessler, B., and Hochreiter, S.
\newblock Gans trained by a two time-scale update rule converge to a local nash
  equilibrium, 2017.

\bibitem[{Hu} et~al.(2014){Hu}, {Lu}, and {Tan}]{face_verfication_inthewild}
{Hu}, J., {Lu}, J., and {Tan}, Y.
\newblock Discriminative deep metric learning for face verification in the
  wild.
\newblock In \emph{2014 IEEE Conference on Computer Vision and Pattern
  Recognition}, 2014.

\bibitem[Huai et~al.(2019)Huai, Xue, Miao, Yao, Su, Chen, and Zhang]{fcn_gen}
Huai, M., Xue, H., Miao, C., Yao, L., Su, L., Chen, C., and Zhang, A.
\newblock Deep metric learning: The generalization analysis and an adaptive
  algorithm.
\newblock In \emph{Proceedings of the Twenty-Eighth International Joint
  Conference on Artificial Intelligence, {IJCAI-19}}, pp.\  2535--2541.
  International Joint Conferences on Artificial Intelligence Organization, 7
  2019.
\newblock \doi{10.24963/ijcai.2019/352}.
\newblock URL \url{https://doi.org/10.24963/ijcai.2019/352}.

\bibitem[Ioffe \& Szegedy(2015)Ioffe and Szegedy]{googlenetv2}
Ioffe, S. and Szegedy, C.
\newblock Batch normalization: Accelerating deep network training by reducing
  internal covariate shift.
\newblock \emph{International Conference on Machine Learning}, 2015.

\bibitem[Jacob et~al.(2019)Jacob, Picard, Histace, and Klein]{horde}
Jacob, P., Picard, D., Histace, A., and Klein, E.
\newblock Metric learning with horde: High-order regularizer for deep
  embeddings.
\newblock In \emph{The IEEE Conference on Computer Vision and Pattern
  Recognition (CVPR)}, 2019.

\bibitem[Jegou et~al.(2011)Jegou, Douze, and Schmid]{recall}
Jegou, H., Douze, M., and Schmid, C.
\newblock Product quantization for nearest neighbor search.
\newblock \emph{IEEE transactions on pattern analysis and machine
  intelligence}, 33\penalty0 (1):\penalty0 117--128, 2011.

\bibitem[Jiang* et~al.(2020)Jiang*, Neyshabur*, Krishnan, Mobahi, and
  Bengio]{generalisation_measures}
Jiang*, Y., Neyshabur*, B., Krishnan, D., Mobahi, H., and Bengio, S.
\newblock Fantastic generalization measures and where to find them.
\newblock In \emph{International Conference on Learning Representations}, 2020.
\newblock URL \url{https://openreview.net/forum?id=SJgIPJBFvH}.

\bibitem[Johnson \& Guestrin(2018)Johnson and Guestrin]{importance_sampling}
Johnson, T.~B. and Guestrin, C.
\newblock Training deep models faster with robust, approximate importance
  sampling.
\newblock In Bengio, S., Wallach, H., Larochelle, H., Grauman, K.,
  Cesa-Bianchi, N., and Garnett, R. (eds.), \emph{Advances in Neural
  Information Processing Systems 31}, pp.\  7265--7275. Curran Associates,
  Inc., 2018.

\bibitem[Keskar et~al.(2016)Keskar, Mudigere, Nocedal, Smelyanskiy, and
  Tang]{keskar2016large}
Keskar, N.~S., Mudigere, D., Nocedal, J., Smelyanskiy, M., and Tang, P. T.~P.
\newblock On large-batch training for deep learning: Generalization gap and
  sharp minima.
\newblock \emph{arXiv preprint arXiv:1609.04836}, 2016.

\bibitem[Kim et~al.(2018)Kim, Goyal, Chawla, Lee, and Kwon]{abe}
Kim, W., Goyal, B., Chawla, K., Lee, J., and Kwon, K.
\newblock Attention-based ensemble for deep metric learning.
\newblock In \emph{Proceedings of the European Conference on Computer Vision
  (ECCV)}, 2018.

\bibitem[Kingma \& Ba(2015)Kingma and Ba]{adam}
Kingma, D.~P. and Ba, J.
\newblock Adam: A method for stochastic optimization.
\newblock 2015.

\bibitem[Krause et~al.(2013)Krause, Stark, Deng, and Fei-Fei]{cars196}
Krause, J., Stark, M., Deng, J., and Fei-Fei, L.
\newblock 3d object representations for fine-grained categorization.
\newblock In \emph{Proceedings of the IEEE International Conference on Computer
  Vision Workshops}, pp.\  554--561, 2013.

\bibitem[Krogh \& Hertz(1992)Krogh and Hertz]{weight_decay}
Krogh, A. and Hertz, J.~A.
\newblock A simple weight decay can improve generalization.
\newblock In \emph{Advances in Neural Information Processing Systems}. 1992.

\bibitem[Lin et~al.(2018)Lin, Duan, Dong, Lu, and Zhou]{dvml}
Lin, X., Duan, Y., Dong, Q., Lu, J., and Zhou, J.
\newblock Deep variational metric learning.
\newblock In \emph{The European Conference on Computer Vision (ECCV)},
  September 2018.

\bibitem[Liu et~al.(2017)Liu, Wen, Yu, Li, Raj, and Song]{sphereface}
Liu, W., Wen, Y., Yu, Z., Li, M., Raj, B., and Song, L.
\newblock Sphereface: Deep hypersphere embedding for face recognition.
\newblock \emph{IEEE Conference on Computer Vision and Pattern Recognition
  (CVPR)}, 2017.

\bibitem[Lloyd(1982)]{kmeans}
Lloyd, S.~P.
\newblock Least squares quantization in pcm.
\newblock \emph{IEEE Trans. Information Theory}, 28:\penalty0 129--136, 1982.

\bibitem[Manning et~al.(2010)Manning, Raghavan, and Sch{\"u}tze]{nmi}
Manning, C., Raghavan, P., and Sch{\"u}tze, H.
\newblock Introduction to information retrieval.
\newblock \emph{Natural Language Engineering}, 16\penalty0 (1):\penalty0
  100--103, 2010.

\bibitem[Milbich et~al.(2020{\natexlab{a}})Milbich, Roth, Bharadhwaj, Sinha,
  Bengio, Ommer, and Cohen]{milbich2020diva}
Milbich, T., Roth, K., Bharadhwaj, H., Sinha, S., Bengio, Y., Ommer, B., and
  Cohen, J.~P.
\newblock Diva: Diverse visual feature aggregation for deep metric learning.
\newblock 2020{\natexlab{a}}.

\bibitem[Milbich et~al.(2020{\natexlab{b}})Milbich, Roth, Brattoli, and
  Ommer]{milbich2020sharing}
Milbich, T., Roth, K., Brattoli, B., and Ommer, B.
\newblock Sharing matters for generalization in deep metric learning,
  2020{\natexlab{b}}.

\bibitem[Mirzasoleiman et~al.(2020)Mirzasoleiman, Bilmes, and
  Leskovec]{coresetoptim}
Mirzasoleiman, B., Bilmes, J., and Leskovec, J.
\newblock Coresets for accelerating incremental gradient methods, 2020.
\newblock URL \url{https://openreview.net/forum?id=SygRikHtvS}.

\bibitem[Misra \& van~der Maaten(2019)Misra and van~der Maaten]{pretextmisra}
Misra, I. and van~der Maaten, L.
\newblock Self-supervised learning of pretext-invariant representations, 2019.

\bibitem[Movshovitz-Attias et~al.(2017)Movshovitz-Attias, Toshev, Leung, Ioffe,
  and Singh]{proxynca}
Movshovitz-Attias, Y., Toshev, A., Leung, T.~K., Ioffe, S., and Singh, S.
\newblock No fuss distance metric learning using proxies.
\newblock In \emph{Proceedings of the IEEE International Conference on Computer
  Vision}, pp.\  360--368, 2017.

\bibitem[Musgrave et~al.(2020)Musgrave, Belongie, and Lim]{musgrave2020metric}
Musgrave, K., Belongie, S., and Lim, S.-N.
\newblock A metric learning reality check, 2020.

\bibitem[Oh~Song et~al.(2016)Oh~Song, Xiang, Jegelka, and Savarese]{lifted}
Oh~Song, H., Xiang, Y., Jegelka, S., and Savarese, S.
\newblock Deep metric learning via lifted structured feature embedding.
\newblock In \emph{Proceedings of the IEEE Conference on Computer Vision and
  Pattern Recognition}, pp.\  4004--4012, 2016.

\bibitem[Opitz et~al.(2018)Opitz, Waltner, Possegger, and Bischof]{abier}
Opitz, M., Waltner, G., Possegger, H., and Bischof, H.
\newblock Deep metric learning with bier: Boosting independent embeddings
  robustly.
\newblock \emph{IEEE transactions on pattern analysis and machine
  intelligence}, 2018.

\bibitem[Paszke et~al.(2017)Paszke, Gross, Chintala, Chanan, Yang, DeVito, Lin,
  Desmaison, Antiga, and Lerer]{pytorch}
Paszke, A., Gross, S., Chintala, S., Chanan, G., Yang, E., DeVito, Z., Lin, Z.,
  Desmaison, A., Antiga, L., and Lerer, A.
\newblock Automatic differentiation in pytorch.
\newblock In \emph{NIPS-W}, 2017.

\bibitem[Qian et~al.(2019)Qian, Shang, Sun, Hu, Li, and Jin]{softriple}
Qian, Q., Shang, L., Sun, B., Hu, J., Li, H., and Jin, R.
\newblock Softtriple loss: Deep metric learning without triplet sampling.
\newblock 2019.

\bibitem[Roth \& Brattoli(2019)Roth and Brattoli]{rothgithub}
Roth, K. and Brattoli, B.
\newblock Deep-metric-learning-baselines.
\newblock \url{https://github.com/Confusezius/Deep-Metric-Learning-Baselines},
  2019.

\bibitem[Roth et~al.(2019)Roth, Brattoli, and Ommer]{mic}
Roth, K., Brattoli, B., and Ommer, B.
\newblock Mic: Mining interclass characteristics for improved metric learning.
\newblock In \emph{Proceedings of the IEEE International Conference on Computer
  Vision}, pp.\  8000--8009, 2019.

\bibitem[Roth et~al.(2020)Roth, Milbich, and Ommer]{roth2020pads}
Roth, K., Milbich, T., and Ommer, B.
\newblock Pads: Policy-adapted sampling for visual similarity learning, 2020.

\bibitem[Rubner et~al.(2000)Rubner, Tomasi, and Guibas]{emd}
Rubner, Y., Tomasi, C., and Guibas, L.~J.
\newblock The earth mover’s distance as a metric for image retrieval.
\newblock \emph{Int. J. Comput. Vision}, 40\penalty0 (2):\penalty0 99–121,
  November 2000.
\newblock ISSN 0920-5691.
\newblock \doi{10.1023/A:1026543900054}.
\newblock URL \url{https://doi.org/10.1023/A:1026543900054}.

\bibitem[Sanakoyeu et~al.(2019)Sanakoyeu, Tschernezki, Buchler, and
  Ommer]{Sanakoyeu_2019_CVPR}
Sanakoyeu, A., Tschernezki, V., Buchler, U., and Ommer, B.
\newblock Divide and conquer the embedding space for metric learning.
\newblock In \emph{The IEEE Conference on Computer Vision and Pattern
  Recognition (CVPR)}, 2019.

\bibitem[Schroff et~al.(2015)Schroff, Kalenichenko, and Philbin]{semihard}
Schroff, F., Kalenichenko, D., and Philbin, J.
\newblock Facenet: A unified embedding for face recognition and clustering.
\newblock In \emph{Proceedings of the IEEE conference on computer vision and
  pattern recognition}, pp.\  815--823, 2015.

\bibitem[Shwartz-Ziv \& Tishby(2017)Shwartz-Ziv and
  Tishby]{shwartzziv2017opening}
Shwartz-Ziv, R. and Tishby, N.
\newblock Opening the black box of deep neural networks via information, 2017.

\bibitem[Sinha et~al.(2019)Sinha, Zhang, Goyal, Bengio, Larochelle, and
  Odena]{sinha2019small}
Sinha, S., Zhang, H., Goyal, A., Bengio, Y., Larochelle, H., and Odena, A.
\newblock Small-gan: Speeding up gan training using core-sets.
\newblock \emph{arXiv preprint arXiv:1910.13540}, 2019.

\bibitem[Smith et~al.(2017)Smith, Kindermans, Ying, and Le]{smith2017don}
Smith, S.~L., Kindermans, P.-J., Ying, C., and Le, Q.~V.
\newblock Don't decay the learning rate, increase the batch size.
\newblock \emph{arXiv preprint arXiv:1711.00489}, 2017.

\bibitem[Sohn(2016)]{npairs}
Sohn, K.
\newblock Improved deep metric learning with multi-class n-pair loss objective.
\newblock In \emph{Advances in Neural Information Processing Systems}, pp.\
  1857--1865, 2016.

\bibitem[Szegedy et~al.(2015)Szegedy, Liu, Jia, Sermanet, Reed, Anguelov,
  Erhan, Vanhoucke, and Rabinovich]{googlenet}
Szegedy, C., Liu, W., Jia, Y., Sermanet, P., Reed, S., Anguelov, D., Erhan, D.,
  Vanhoucke, V., and Rabinovich, A.
\newblock Going deeper with convolutions.
\newblock In \emph{Proceedings of the IEEE conference on computer vision and
  pattern recognition}, pp.\  1--9, 2015.

\bibitem[Tishby \& Zaslavsky(2015)Tishby and Zaslavsky]{tishby2015deep}
Tishby, N. and Zaslavsky, N.
\newblock Deep learning and the information bottleneck principle, 2015.

\bibitem[Ustinova \& Lempitsky(2016)Ustinova and Lempitsky]{histogram}
Ustinova, E. and Lempitsky, V.
\newblock Learning deep embeddings with histogram loss.
\newblock In \emph{Advances in Neural Information Processing Systems}, 2016.

\bibitem[Verma et~al.(2018)Verma, Lamb, Beckham, Najafi, Mitliagkas, Courville,
  Lopez-Paz, and Bengio]{manifoldmixup}
Verma, V., Lamb, A., Beckham, C., Najafi, A., Mitliagkas, I., Courville, A.,
  Lopez-Paz, D., and Bengio, Y.
\newblock Manifold mixup: Better representations by interpolating hidden
  states, 2018.

\bibitem[Wah et~al.(2011)Wah, Branson, Welinder, Perona, and
  Belongie]{cub200-2011}
Wah, C., Branson, S., Welinder, P., Perona, P., and Belongie, S.
\newblock The caltech-ucsd birds-200-2011 dataset.
\newblock 2011.

\bibitem[Wang et~al.(2017)Wang, Zhou, Wen, Liu, and Lin]{angular}
Wang, J., Zhou, F., Wen, S., Liu, X., and Lin, Y.
\newblock Deep metric learning with angular loss.
\newblock In \emph{Proceedings of the IEEE International Conference on Computer
  Vision}, pp.\  2593--2601, 2017.

\bibitem[Wang et~al.(2019{\natexlab{a}})Wang, Han, Huang, Dong, and
  Scott]{multisimilarity}
Wang, X., Han, X., Huang, W., Dong, D., and Scott, M.~R.
\newblock Multi-similarity loss with general pair weighting for deep metric
  learning, 2019{\natexlab{a}}.

\bibitem[Wang et~al.(2019{\natexlab{b}})Wang, Hua, Kodirov, Hu, Garnier, and
  Robertson]{rankedlist}
Wang, X., Hua, Y., Kodirov, E., Hu, G., Garnier, R., and Robertson, N.~M.
\newblock Ranked list loss for deep metric learning.
\newblock \emph{The IEEE Conference on Computer Vision and Pattern Recognition
  (CVPR)}, 2019{\natexlab{b}}.

\bibitem[Wu et~al.(2017)Wu, Manmatha, Smola, and Krahenbuhl]{margin}
Wu, C.-Y., Manmatha, R., Smola, A.~J., and Krahenbuhl, P.
\newblock Sampling matters in deep embedding learning.
\newblock In \emph{Proceedings of the IEEE International Conference on Computer
  Vision}, pp.\  2840--2848, 2017.

\bibitem[Xuan et~al.(2018)Xuan, Souvenir, and Pless]{dreml}
Xuan, H., Souvenir, R., and Pless, R.
\newblock Deep randomized ensembles for metric learning.
\newblock In \emph{Proceedings of the European Conference on Computer Vision
  (ECCV)}, pp.\  723--734, 2018.

\bibitem[Yu et~al.(2018)Yu, Liu, Gong, Ding, and Tao]{yu2018correcting}
Yu, B., Liu, T., Gong, M., Ding, C., and Tao, D.
\newblock Correcting the triplet selection bias for triplet loss.
\newblock In \emph{Proceedings of the European Conference on Computer Vision
  (ECCV)}, pp.\  71--87, 2018.

\bibitem[Yuan et~al.(2019)Yuan, Deng, Tang, Tang, and Chen]{signal2noise}
Yuan, T., Deng, W., Tang, J., Tang, Y., and Chen, B.
\newblock Signal-to-noise ratio: A robust distance metric for deep metric
  learning.
\newblock \emph{The IEEE Conference on Computer Vision and Pattern Recognition
  (CVPR)}, 2019.

\bibitem[Zhai \& Wu(2018)Zhai and Wu]{zhai2018classification}
Zhai, A. and Wu, H.-Y.
\newblock Classification is a strong baseline for deep metric learning, 2018.

\bibitem[Zheng et~al.(2019)Zheng, Chen, Lu, and Zhou]{hardness-aware}
Zheng, W., Chen, Z., Lu, J., and Zhou, J.
\newblock Hardness-aware deep metric learning.
\newblock \emph{The IEEE Conference on Computer Vision and Pattern Recognition
  (CVPR)}, 2019.

\end{thebibliography}
